\PassOptionsToPackage{numbers, compress}{natbib}
\documentclass{article}

 \usepackage[preprint]{neurips_2026}

\usepackage[utf8]{inputenc} 
\usepackage[T1]{fontenc}    
\usepackage{hyperref}       
\usepackage{url}            
\usepackage{booktabs}       
\usepackage{amsfonts}       
\usepackage{nicefrac}       
\usepackage{microtype}      
\usepackage{xcolor}         

\usepackage[table]{xcolor}       
\usepackage{enumitem}
\usepackage{amsmath}
\usepackage{tabularx}
\usepackage{boldline}
\usepackage{booktabs}
\usepackage{arydshln}
\usepackage{multirow}
\usepackage{colortbl}
\usepackage{svg}
\usepackage{tabularx}
\usepackage{makecell}
\usepackage{listings}

\usepackage{booktabs}
\usepackage{multirow}
\usepackage{tabularx}
\usepackage{makecell}
\usepackage[table]{xcolor} %
\usepackage{array}
\usepackage{tcolorbox}
\usepackage{xspace}
\usepackage{wrapfig}
\usepackage{caption}
\usepackage{subcaption}
\title{SVoT: State-aware Visualization-of-Thought for Spatial Reasoning via
Reinforcement Learning}

\author{
Chao Lei \quad
Yanbei Jiang \quad
Markus Hiller \quad
Zhijian Zhou \\
\textbf{Xunye Tian \quad
Krista A. Ehinger \quad
Nir Lipovetzky} \\
School of Computing and Information Systems \\
The University of Melbourne \\
Melbourne, Australia \\
\texttt{\{clei1,yanbeij\}@student.unimelb.edu.au} \\
\texttt{\{zhijianzhou.ml, xunyetian.ml\}@gmail.com} \\
\texttt{\{m.hiller, nir.lipovetzky, kris.ehinger\}@unimelb.edu.au}\\
}

\begin{document}

\maketitle

\begin{abstract}
Spatial reasoning remains a challenge for Multimodal Large Language Models (MLLMs), as it requires reliable multi-hop inference over both intermediate states and state transitions. Current studies often leave intermediate states unverified and treat state transitions as implicit processes, which limits reliability in multi-hop spatial reasoning. To address this, we propose State-aware Visualization-of-Thought (SVoT), a reinforcement learning framework that generates interleaved, verifiable intermediate states and visualizations. SVoT integrates transition reasoning chains into the generation processes, enabling the model to verify action preconditions and effects through interleaved textual and visual reasoning. We train SVoT via Group Relative Policy Optimization (GRPO), instantiating verification through reward design and evaluating the efficacy of different fine-grained rewards. As existing benchmarks reduce state transitions to single-variable updates, substantially simplifying the problems, we establish five domains by extending classical environments and introducing two novel domains, \textsc{Pacman} and \textsc{Gather}, that require multi-object interactions and numerical reasoning. These domains support systematic evaluation of multi-hop spatial reasoning with quantitative verification of generated intermediate states and transition reasoning. SVoT with transition-aware supervision achieves state-of-the-art performance across the introduced domains, yielding up to a 65\% absolute accuracy gain on out-of-distribution test sets.
\end{abstract}

\section{Introduction}

Planning constitutes a core reasoning capability in Multimodal Large Language Models (MLLMs) \cite{wu2025vsp}. A prominent evaluation setting for planning in MLLMs is spatial reasoning \cite{zhang2025mllms}, a domain essential to navigation \cite{qiao2025navbench,xu2025flame}, robotics \cite{hu2023look}, and autonomous driving \cite{tian2024drivevlm,ma2024dolphins}. Spatial reasoning entails the ability to perceive and analyze object relationships, state transitions, and environmental interactions \cite{wu2024mind}. Although existing Visual Question Answering (VQA) benchmarks~\cite{chen2024spatialvlm,liu2023visual,wang2024picture,li2024topviewrs} have been proposed to evaluate  spatial reasoning in MLLMs, they primarily emphasize static spatial relations, e.g., relative object positions, with limited coverage of planning-oriented capabilities. In contrast, multi-hop spatial reasoning requires sequential state tracking, i.e., accurate next-state predictions under a given action sequence \cite{wu2024mind,li2025imagine}, which is central to planning. This dynamic nature compels the model to consistently identify objects, understand temporal evolution, and infer action-conditioned state transitions.

Grid-based environments provide a natural setting for evaluating  sequential state tracking. They define discrete state variables (e.g., coordinates) and deterministic  transition dynamics, which facilitate rigorous verification of state updates and validity constraints. Recent studies \cite{wu2024mind,li2025imagine} evaluate sequential state tracking within grid-world benchmarks such as \textsc{Maze} \cite{ivanitskiy2023configurable}, \textsc{FrozenLake} \cite{brockman2016openai}, and \textsc{Polyomino Tilings} \cite{golomb1966tiling}.  Moreover, grid-based representations are widely used in real-world spatial applications, where continuous spaces are discretized into occupancy grids or cost maps for tractable collision checking, path planning, and multi-agent coordination in domains such as autonomous driving, robot navigation, and warehouse robotics~\cite{mohajerin2019multi,liu2023hybrid,stern2019multi,salzman2020research}. Critically, accurate state tracking is essential for safe planning in these real-world applications. Thus, improving the reliability and verifiability of sequential state tracking in grid-based environments addresses a fundamental requirement for applying MLLMs to real-world spatial reasoning tasks.

\begin{figure*}
    \centering
    \includegraphics[width=1\linewidth]{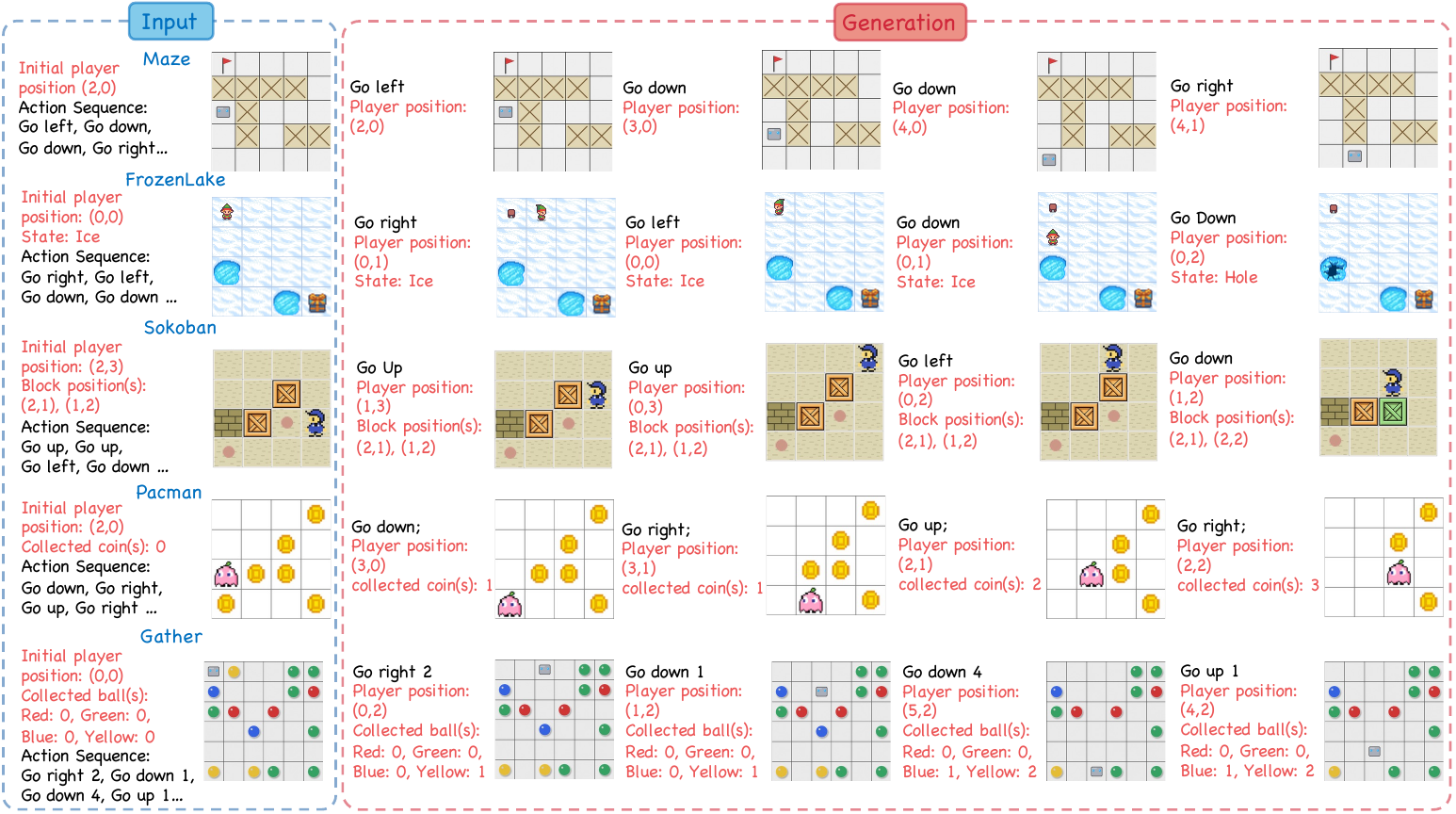}
    \vspace{-0.6cm}
    \caption{Illustration of the five domains used in SVoT. Coordinates are (row, column), starting from (0,0) at the top-left corner. For conciseness, we omit the domain description from the input.}
        
    \label{fig:domains}
    \vspace{-0.5cm}
\end{figure*}

Benefiting from the faculties of constructing mental images and performing mental simulation, humans demonstrate robust spatial awareness \cite{shepard1971mental,tolman1948cognitive}. Inspired by this, \citet{wu2024mind} proposed Visualization-of-Thought (VoT), which constructs a textual visualization after each reasoning step to explicitly track intermediate states. Notably, VoT demonstrates that accurate sequential state tracking is essential for multi-hop spatial reasoning. Advancing this paradigm, Multimodal Visualization-of-Thought (MVoT) \cite{li2025imagine} identifies native multimodal generation as the promising pathway for implementing sequential state tracking. By leveraging a multimodal-native backbone to generate actual visualizations, MVoT unifies text and vision within the reasoning traces, offering a visually grounded reasoning process. MVoT achieves state-of-the-art performance in sequential state tracking across multiple grid-based domains.

Despite this progress, VoT and MVoT suffer from two critical limitations. First, existing evaluations typically restrict goal specifications to a single state variable or a binary success flag. This reduces each state transition to a single-variable update and weakens the multi-object interactions that make multi-hop spatial reasoning genuinely challenging. Second, the absence of intermediate-state verification and explicit state-transition reasoning prevents reliable assessment of whether the model reasons correctly at each step or merely reaches the correct final state by chance, thereby limiting the reliability of sequential state tracking.

To address both limitations, we propose State-aware Visualization-of-Thought (SVoT), a reinforcement learning (RL)-based framework that improves MLLMs' sequential state tracking through explicit intermediate-state verification and state-transition reasoning. We further evaluate SVoT in richer dynamic environments beyond single-object interactions. In contrast to MVoT, which relies mainly on visualizations as the reasoning trace, SVoT augments the trace with an explicit, structured representation for each intermediate state to support verification and integrates transition reasoning chains into the generation of both intermediate states and visualizations. Transition reasoning chains enable the model to explicitly verify action preconditions and effects through interleaved textual and visual reasoning, achieving both reliable next-state prediction and faithful visual synthesis. We further examine reward design as the verification mechanism in SVoT, adopting both an Outcome Reward Model (ORM) that supervises only the state description and visualization, and a fine-grained Process Reward Model (PRM) that additionally optimizes the reasoning trajectory.

To support rigorous evaluation of  richer dynamics, we establish five grid-based domains, extending classical environments such as \textsc{Maze} and \textsc{FrozenLake}, incorporating \textsc{Sokoban}, and introducing the novel \textsc{Pacman} and \textsc{Gather} domains, as illustrated in Figure \ref{fig:domains}. These tasks feature multiple interacting objects (e.g., walls, holes, blocks, coins, balls) with diverse attributes and constraints (e.g., collisions, hazards, positions, counts, colors), requiring explicit transition reasoning to support non-trivial numerical and causal inference. Their discrete, object-centric formulation provides structured representations of states and state transitions, allowing quantitative verification of both state-representation and transition-reasoning correctness. Detailed related work is provided in Appendix~\ref{Related Work}.

We outline our contributions as follows: 1) We propose SVoT, a RL-based framework that unifies verifiable intermediate state with visual generation and integrates transition reasoning chains as explicit reasoning steps to enhance MLLMs' sequential state tracking in grid-based environments. 2) We introduce five grid-based domains that enable systematic evaluation of multi-hop spatial reasoning and provide explicit representations of intermediate states and state transitions, supporting quantitative verification of intermediate-state and transition-reasoning correctness. 3) We conduct extensive experiments across the introduced domains, demonstrating that SVoT achieves state-of-the-art performance in sequential state tracking, with up to a 65\% absolute improvement on out-of-distribution test sets. We further analyze the impact of alternative reward models, and highlight failure cases and remaining challenges in multi-hop spatial reasoning.

\section{State-aware Visualization-of-Thought}

In this work, we study sequential state tracking for multi-hop spatial reasoning. We formulate the task as $\mathcal{P}=\langle \mathcal{X}, g \rangle$, where $\mathcal{X}=\langle d, v_0, \mathcal{A} \rangle$ denotes the task specification, comprising the domain description $d$, the initial-state visualization $v_0$, and the action sequence $\mathcal{A}=\{a_i\}_{i=1}^{n}$, and $g$ is the ground-truth goal configuration. Given $\mathcal{X}$,  a pre-trained MLLM with parameters $\theta$, $\mathcal{M}_{\theta}$, is expected to generate a sequence of intermediate states $\mathcal{Z}=\{z_i\}_{i=1}^{n}$ and then predict the goal configuration $g$. Each $z_i$ is generated autoregressively conditioned on $\mathcal{X}$ and the preceding states $z_{<i}$, while $g$ is predicted conditioned on $\mathcal{X}$ and the complete trajectory $\mathcal{Z}$:
{
\setlength{\abovedisplayskip}{6pt}
\setlength{\belowdisplayskip}{6pt}
\begin{align}
z_i &\sim \mathcal{M}_{\theta}(\cdot \mid \mathcal{X}, \{z_j\}_{j=1}^{i-1}), \\
g &\sim \mathcal{M}_{\theta}(\cdot \mid \mathcal{X}, \{z_j\}_{j=1}^{n}).
\end{align}
}

\setlength{\intextsep}{0pt} %
\begin{wrapfigure}{R}{0.63\textwidth}
    \centering
        \setlength{\abovecaptionskip}{1pt}
    \includegraphics[width=0.63\textwidth]{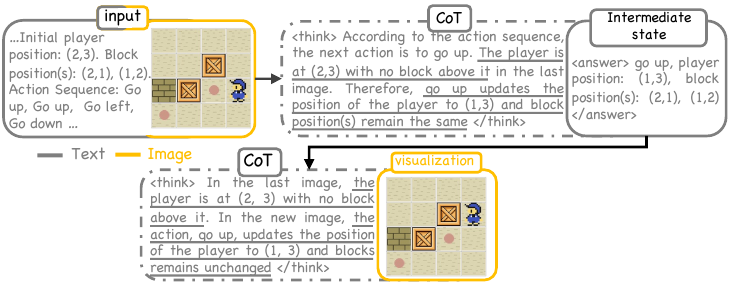} 
    \caption{Examples of the CoT (transition reasoning chain) in SVoT used to guide the generation of intermediate state and visualization for the \textsc{Sokoban} problem in Figure~\ref{fig:domains}}
    \label{fig:CoT_Example}
\end{wrapfigure}

Although both VoT and MVoT emphasize the significance of intermediate state $z_i$ for sequential state tracking, the  definition of $z_i$ remains underspecified. In tasks such as \textsc{Maze} and \textsc{FrozenLake}, MVoT typically reduces $z_i$ to merely the next action prediction, providing sparse information for visualization and subsequent state prediction. In contrast, SVoT formally defines the intermediate state as a tuple $z_i=\langle a_i, s_i \rangle$, containing the action $a_i$ and the state description $s_i$, resulting from applying $a_i$ to the previous state $s_{i-1}$. In SVoT, we generate the ground truth for each $s_i$ by anchoring the process with an initial state $s_0$. Consequently, each $s_i$ is derived via a deterministic transition function $f$, $s_{i}=f(s_{i-1}, a_i)$, based on the specified action preconditions and effects. For each domain, Figure~\ref{fig:domains} illustrates the intermediate state $z_i$ (comprising the ground truth $a_i$ and $s_i$), and the initial state description $s_0$ in the input, highlighted in red. The state descriptions $s_i$ in SVoT capture precise object layout, relationships, and attributes to guide subsequent generations and enable the quantitative evaluation of $s_i$, e.g., verifying the player's coordinate (e.g., $(0,2)$) or the exact number of collected coins (e.g., $2$).

Central to SVoT is the transition reasoning chain  $c_i$, which acts as the transition function, helping the model derive the intermediate state  $z_i$ and the visualization $v_i$ by verifying action preconditions and applying action effects, closely aligning with classical planning assumptions.  Unlike standard Chain-of-Thought (CoT) in LLMs, $c_i$ encodes both semantic and spatial information (e.g., ``the player is at (2,3) with no block above it''), as illustrated in Figure \ref{fig:CoT_Example}. This enables SVoT to move beyond the implicit state updates of MVoT toward a structured and verifiable reasoning process. By grounding each generation in a precise state-updating process, $c_i$ bridges abstract spatial reasoning and concrete multimodal generation, ensuring that state descriptions and visualizations remain consistent with the underlying dynamics. For notational clarity, we use $c_i^z$ and $c_i^v$ to denote the transition reasoning chains for $z_i$ and $v_i$, respectively, while using $c_i$ as the general form (or simply ``CoT'' for brevity).

\setlength{\intextsep}{0pt} %
\begin{wrapfigure}{R}{0.54\textwidth}
    \centering
        \setlength{\abovecaptionskip}{1pt}
    \includegraphics[width=0.54\textwidth]{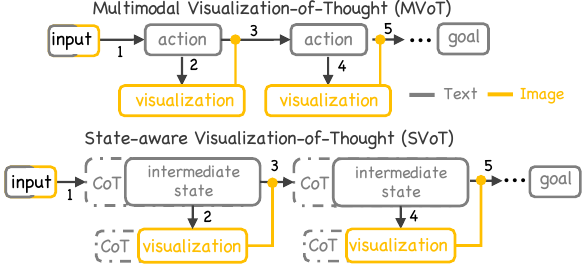}
    \caption{The architectures of MVoT and SVoT.} 
    \label{fig:flowchart}
    \vspace{0.1cm}
\end{wrapfigure}

To unify reasoning and generation for sequential state tracking, SVoT augments the task specification $\mathcal{X}$ with an initial state description $s_0$, resulting in $\mathcal{X}=\langle d, s_0, v_0, \mathcal{A} \rangle$. SVoT represents each intermediate state as $z_i=\langle a_i, s_i \rangle$. In addition, SVoT produces a corresponding visualization $v_i$ for each $z_i$. Crucially, the MLLM is instructed to generate a transition reasoning chain $c_i$ in conjunction with $z_i$ and $v_i$, yielding an interleaved multimodal generation process that naturally supports CoT reasoning. In SVoT, $c_i$, $z_{i}$, and $v_i$ are predicted conditioned on $\mathcal{X}$ and the preceding generated states and visualizations, and the goal configuration $g$ is then predicted from the complete multimodal trajectory. To satisfy context-window constraints while preserving the explicit trajectory, we exclude previously generated transition reasoning chains from the conditioning context in each generation. Equations~\eqref{eq:svot_state}--\eqref{eq:svot_goal} formalize the SVoT generation process:
{
\setlength{\abovedisplayskip}{6pt}
\setlength{\belowdisplayskip}{6pt}
\begin{align}
c_{i}^z, z_{i} &\sim \mathcal{M}_{\theta}( \cdot \mid \mathcal{X}, \{z_j, v_j\}_{j=1}^{i-1}), \label{eq:svot_state} \\
c_i^v, v_i &\sim \mathcal{M}_{\theta}(\cdot \mid \mathcal{X}, \{z_j, v_j\}_{j=1}^{i-1}, z_i), \label{eq:svot_vis} \\
g &\sim \mathcal{M}_{\theta}(\cdot \mid \mathcal{X}, \{z_j, v_j\}_{j=1}^{n}). \label{eq:svot_goal}
\end{align}
}

Figure~\ref{fig:flowchart} illustrates the architectural differences between SVoT and MVoT. In MVoT, the intermediate state $z_i$ is typically reduced to an isolated action, e.g., $z_i=a_i$, without explicit transition reasoning to support subsequent state or visualization generation. For conciseness, we make no distinction between goal prediction and state prediction in SVoT unless explicitly noted.

\section{Training Strategy} \label{Training}

SVoT employs Anole-7B~\cite{chern2024anole} as the backbone model, a unified autoregressive MLLM that encodes visual and textual inputs as a single stream of discrete tokens. This architecture supports joint text-and-visual generation within a unified autoregressive sequence, enabling end-to-end optimization. Anole builds on Chameleon \cite{team2024chameleon} without explicitly optimizing CoT capabilities. We implement a two-stage training pipeline to develop CoT-based multi-hop spatial reasoning capabilities for this backbone.  In both stages, each training prompt is constructed with the ground-truth multimodal  trajectory prefix. In the first stage, we perform Supervised Fine-Tuning (SFT) to train the model to generate transition-reasoning-guided state outputs (e.g., $\{c_i^z, z_i\}$) and  visual counterparts (e.g., $\{c_i^v, v_i\}$), utilizing the loss function introduced in MVoT (details in Appendix \ref{MVoT}).  
This stage establishes a robust initialization for the subsequent RL phase. In the second stage, we employ Group Relative Policy Optimization (GRPO) \cite{shao2024deepseekmath} to encourage  exploration of flexible and effective transition reasoning chains for synthesizing $z_i$ and $v_i$.

SVoT defines three reward functions for GRPO: (1) a state reward $r_z$ for the correctness of each intermediate state $z_i$, (2) a visual reward $r_v$ for the fidelity of each visualization $v_i$, and (3) a reasoning reward $r_c$ for the faithfulness of each transition reasoning chain $c_i$. For the text modality rewards, $r_z$ and $r_c$, SVoT parses the prediction $\hat{y}$ and the ground truth $y$ into a set of semantic elements $P(y)$ and defines $r_z$  and $r_c$ as the average exact-match accuracy across all fields $p \in P(y)$, as formulated in Equation~\eqref{eq:unified_reward}:
{ 
\begin{equation}
    r(\hat{y}, y) = \frac{1}{|P(y)|} \sum_{p \in P(y)} \mathbb{I}\left[p(\hat{y})=p(y)\right].
    \label{eq:unified_reward}
\end{equation}
}

Here, $\mathbb{I}[\cdot]$ denotes the indicator function. For $r_z$ (where $y=z_i$), $P(z_i)$ consists of discrete fields derived from the action and state description (e.g., action direction, player coordinates, and the number of collected coins in \textsc{Pacman}). For $r_c$ (where $y=c_i$), $P(c_i)$ contains keywords extracted from action preconditions and effects (e.g., ``with no block above it'' for preconditions, and ``block position(s) remain the same'' for effects in \textsc{Sokoban} as shown in Figure \ref{fig:CoT_Example}).

For the visual reward $r_v$, we partition the generated image $\hat{v}$ and the corresponding ground-truth image $v$ into aligned $n \times n$ grids of cells, denoted by $\hat{v}_{ij}$ and $v_{ij}$, and compute a weighted cell-wise matching score that measures local visual consistency. Since reliable semantic labels are often unavailable, we employ a variance-based heuristic for automatic foreground detection, where a grid cell is classified as foreground if its pixel-wise color variance exceeds a threshold. This allows the $r_v$ to assign higher weights to semantically salient regions (e.g., players and boxes) than to the background, where the heuristic tends to highlight texture-rich objects and treat relatively uniform cells as background. The visual reward $r_v$ combines a cell-level matching term $r_{\mathrm{str}}$ with an image-quality term $r_{\mathrm{qua}}$:
{
\setlength{\abovedisplayskip}{6pt}
\setlength{\belowdisplayskip}{6pt}
\begin{align}
    r_v(\hat{v}, v) &= r_{\mathrm{str}}(\hat{v}, v) \cdot r_{\mathrm{qua}}(\hat{v}, v), \label{eq:image_reward} \\
    r_{\mathrm{str}}(\hat{v}, v) &= \sum_{(i,j)} w_{ij} \cdot \mathbb{I}\left[ \mathcal{S}(\hat{v}_{ij}, v_{ij}) \ge \tau \right], \label{eq:image_reward_str} \\
    r_{\mathrm{qua}}(\hat{v}, v) &= \mathrm{clip}\left( \frac{\phi(\hat{v}; \Omega)}{\phi(v; \Omega) + \epsilon}, 0, 1 \right). \label{eq:image_reward_qua}
\end{align}
}
\par\noindent Here, $r_{\mathrm{str}}$ measures cell-level consistency using structural, edge, and color features, while $r_{\mathrm{qua}}$ penalizes blurry predictions by measuring foreground sharpness. In Eq.~\eqref{eq:image_reward_str}, a cell contributes to the reward only when its composite similarity score $\mathcal{S}$  exceeds the threshold $\tau$. The weights $w_{ij}$ are normalized such that $\sum w_{ij} = 1$ and assign higher weight to foreground cells than background cells to encourage semantic focus. $\mathcal{S}$ combines three features:
{\setlength{\abovedisplayskip}{6pt}\setlength{\belowdisplayskip}{6pt}
\begin{equation}
\mathcal{S}(\cdot) = \lambda_h s_{\mathrm{hash}}(\cdot) + \lambda_e s_{\mathrm{edge}}(\cdot) + \lambda_c s_{\mathrm{color}}(\cdot),
\end{equation}}
\par\noindent where $s_{\mathrm{hash}}$ measures perceptual-hash similarity to capture global structure, $s_{\mathrm{edge}}$ computes edge-map IoU using Sobel operators to reflect local texture boundaries, and $s_{\mathrm{color}}$ evaluates joint RGB histogram correlation to assess color consistency. In Eq.~\eqref{eq:image_reward_qua}, $\Omega$ denotes the foreground cells, and $\phi(\cdot; \Omega)$ calculates the average Laplacian variance within $\Omega$ to measure sharpness. Detailed visual reward formulation and hyperparameter settings are available in Appendices~\ref{Detailed Formulation of Visual Reward} and~\ref{Hyperparameters Analysis in Visual Reward}.

For each prompt, GRPO samples a group of $G$ completions $\{o_g\}_{g=1}^{G}$. In SVoT, completions within a group belong to the same output type, either $\{c^z, z\}$ or $\{c^v, v\}$. The reward $r_g$ for completion $o_g$ is computed as a weighted sum of individual rewards (e.g., $r_g=0.5\,r_c+0.5\,r_z$).  
GRPO then normalizes rewards within each group to compute relative advantages, promoting higher-reward completions while suppressing suboptimal ones. SVoT optimizes the policy using the standard GRPO objective. Full details of GRPO in SVoT are available in Appendix~\ref{GRPO}.

\begin{figure*}
    \centering
    \includegraphics[width=1\linewidth]{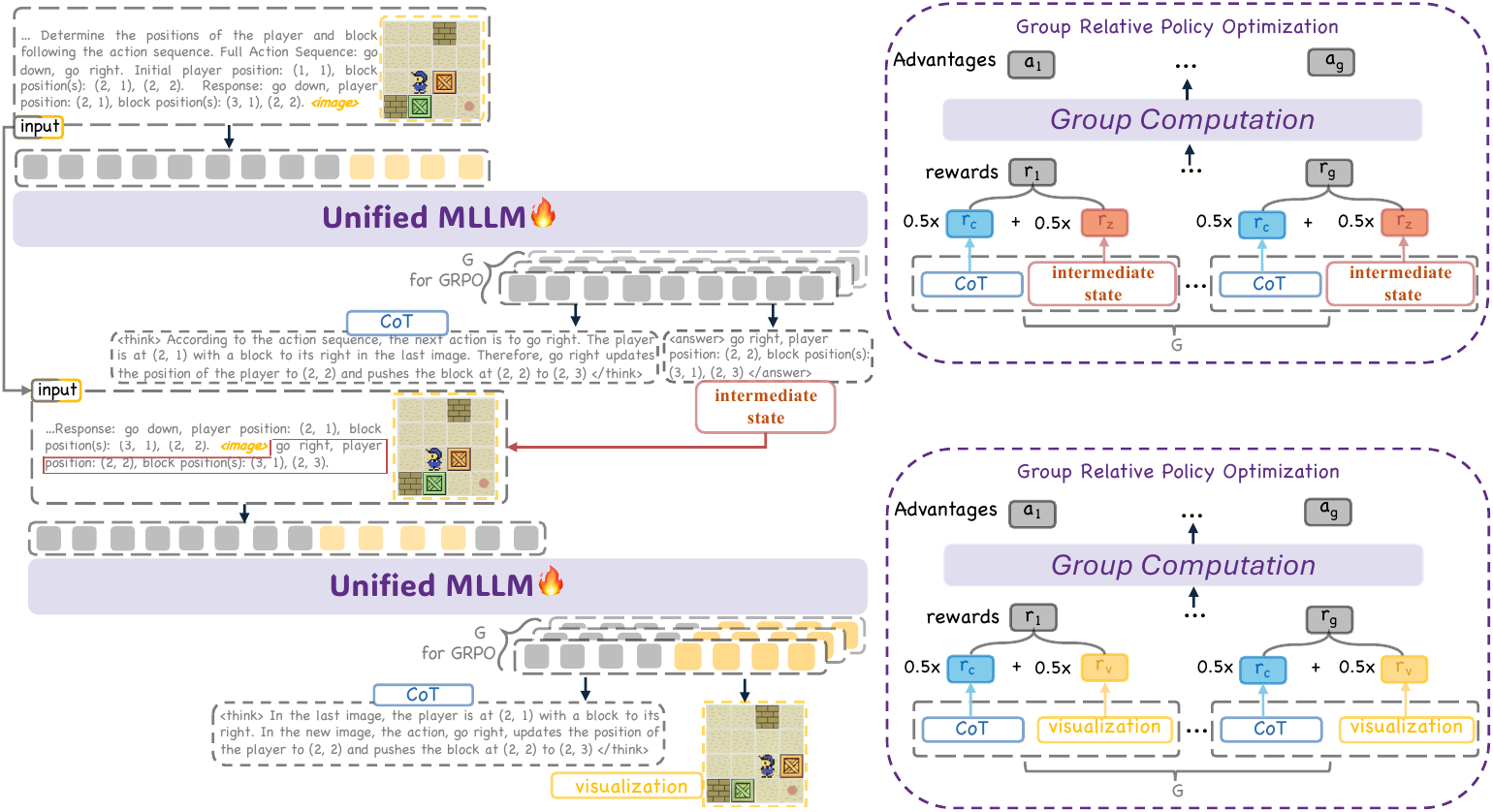}
    \vspace{-0.4cm}
    \caption{GRPO training in SVoT with PRM. Left: sampling CoT with intermediate state and CoT with visualization completions. Right: reward and advantage computation for GRPO updates. }
    \vspace{-0.6cm}
    
    \label{fig:grpo}
\end{figure*}

In SVoT, the reasoning reward $r_c$ evaluates the fidelity of the transition reasoning chain against its ground truth, ensuring that the model's logic is sound and faithful rather than solely optimized for the output, an approach aligning with a Process Reward Model (PRM) \cite{lightman2024lets,uesato2022solving}. In contrast, we explore a simplified $r_g$ that retains only $r_z$ or $r_v$, while removing reasoning reward $r_c$ akin to an Outcome Reward Model (ORM) \cite{uesato2022solving}.  We empirically investigate whether explicit supervision of the transition reasoning process via PRM yields superior sequential state tracking performance compared to sparse, outcome-only signals used by ORM. Figure~\ref{fig:grpo} illustrates the GRPO pipeline in SVoT with PRM.

\section{Spatial Reasoning Tasks}
To support rigorous evaluation of multi-hop spatial reasoning, we establish five grid-based domains (Figure \ref{fig:domains}) of varying complexity, where the goal configuration $g$ is defined as the final state description induced by the given action sequence $\mathcal{A}$. Actions are single-cell moves in most domains, and multi-step moves in \textsc{Gather}, restricted to the four cardinal directions (up/down/left/right), with domain-specific preconditions and effects. Brief introductions to each domain are provided below, with full construction details and descriptions available in Appendix~\ref{Domain Generation}.

\begin{itemize}[noitemsep, topsep=0pt, leftmargin=*]

\item \textsc{Maze}: Building on the \textsc{Maze} environment from MVoT, we additionally include invalid movements that require models to infer action effects, where boundary violations and wall collisions leave the agent's position unchanged. The objective is to predict the agent's final coordinate.

\item \textsc{FrozenLake}: We adopt \textsc{FrozenLake} from MVoT, where tile distinctions (e.g., ice vs.\ holes vs.\ goal) demand precise perception. The task is to predict the final player coordinate and tile type.

\item \textsc{Sokoban}: A classical planning puzzle with multi-object interactions, where player moves can push a block directly in front, resulting in block displacements. The task is to predict the final coordinates of the player and all blocks.

\item \textsc{Pacman}: The player navigates a grid to collect coins, integrating numerical reasoning into sequential state tracking under multi-object interactions. The task requires predicting the cumulative number of collected coins over the action sequence and the agent's final coordinate.

\item \textsc{Gather}: An advanced variant of \textsc{Pacman}, introducing  multi-step actions (e.g., ``move down 4 steps”) and color-specific objectives, where each movement collects colored balls distributed along the trajectory. The task requires predicting the number of colored balls collected and the agent's final position. This domain presents a substantial challenge, demanding long-horizon state tracking, numerical reasoning, and visual grounding.

\end{itemize}

Across all domains, we vary grid sizes from $4$ to $6$ to maintain image quality, while extending to $5$ to $7$ for \textsc{Maze} to ensure feasible paths. For each size, we scale the action sequence lengths to evaluate long-horizon state tracking, and adjust the number of interactive objects to ensure diversity. For each domain and grid size, we construct 500 deduplicated training instances, comprising 100 valid maps, each paired with 5 valid action sequences. To assess generalization, we define In-Distribution (ID) and Out-of-Distribution (OOD) test settings. ID instances match the training distribution, while OOD instances introduce longer action sequences and larger numbers of interactive objects, testing robustness to configurations unseen during training.  We generate separate ID and OOD test sets by sampling 60 unseen maps per domain and grid size and pairing each map with 2 valid action sequences under the corresponding configuration, yielding 120 test instances per domain–size pair in each setting. See Appendix~\ref{Dataset Collection} for dataset collection details and domain statistics.

\definecolor{bestgray}{gray}{0.92} 
\definecolor{linegray}{gray}{0.65} 
\newcommand{\blank}{}

\begin{table*}[t]
\centering
\scriptsize
\setlength{\tabcolsep}{1.8pt}
\renewcommand{\arraystretch}{1}

\begin{tabular}{@{} c c l *{4}{c} !{\color{linegray}\vrule} *{4}{c} !{\color{linegray}\vrule} *{4}{c} !{\color{linegray}\vrule} *{4}{c} !{\color{linegray}\vrule} *{4}{c} @{}}
\toprule
\multirow{3}{*}{\textbf{Size}} 
& \multirow{3}{*}{\textbf{Backbone}}
& \multirow{3}{*}{\textbf{Method}}
& \multicolumn{4}{c}{\textbf{Maze}}
& \multicolumn{4}{c}{\textbf{FrozenLake}}
& \multicolumn{4}{c}{\textbf{Sokoban}}
& \multicolumn{4}{c}{\textbf{Pacman}}
& \multicolumn{4}{c}{\textbf{Gather}} \\
\cmidrule(lr){4-7}\cmidrule(lr){8-11}\cmidrule(lr){12-15}\cmidrule(lr){16-19}\cmidrule(lr){20-23}
& & 
& \multicolumn{2}{c}{ID} & \multicolumn{2}{c}{OOD}
& \multicolumn{2}{c}{ID} & \multicolumn{2}{c}{OOD}
& \multicolumn{2}{c}{ID} & \multicolumn{2}{c}{OOD}
& \multicolumn{2}{c}{ID} & \multicolumn{2}{c}{OOD}
& \multicolumn{2}{c}{ID} & \multicolumn{2}{c}{OOD} \\
\cmidrule(lr){4-5}\cmidrule(lr){6-7}
\cmidrule(lr){8-9}\cmidrule(lr){10-11}
\cmidrule(lr){12-13}\cmidrule(lr){14-15}
\cmidrule(lr){16-17}\cmidrule(lr){18-19}
\cmidrule(lr){20-21}\cmidrule(lr){22-23}
& & 
& C & F & C & F
& C & F & C & F
& C & F & C & F
& C & F & C & F
& C & F & C & F \\
\midrule

\multirow{5}{*}{\textbf{4}}
& GPT-4o 
& T-CoT & \blank & \blank & \blank & \blank    
& 100 & 83.3 & 90.0 & 56.7    
& 63.3 & 48.3 & 56.7 & 36.7    
& 90.0 & 26.7 & 83.3 & 10.0    
& 53.8 & 23.3 & 46.7 & 13.3 \\
\cmidrule(lr){2-23}
& \multirow{4}{*}{Anole}
& T-CoT       & - & - & - & -    
& 75.0 & 33.3 & 66.7 & 33.3   
& 65.0 & 5.0  & 53.3 & 0.0    
& 71.7 & 53.3 & 50.0 & 36.7    
& 41.7 & 6.7  & 36.7 & 0.0 \\
&
& MVoT      & - & - & - & -    
& 86.7 & 46.7 & 80.0 & 56.7   
& 73.3 & 15.0 & 61.7 & 3.3    
& 73.3 & 56.7 & 63.3 & 46.7    
& 46.7 & 13.3 & 36.7 & 3.3 \\
&
& SVoT$_{\text{o}}$ & - & - & - & -    
& \textbf{100} & \textbf{86.7} & \textbf{88.3} & 63.3   
& 73.3 & 71.7 & 70.0 & 50.0    
& \textbf{100} & 63.3 & 86.7 & 61.7    
& \textbf{60.0} & 40.0 & 43.3 & 10.0 \\
\rowcolor{bestgray}
&
& SVoT$_{\text{p}}$ & - & - & - & -    
& \textbf{100} & 83.3 & \textbf{88.3} & \textbf{68.3}   
& \textbf{86.7}  & \textbf{78.3} & \textbf{80.0} & \textbf{68.3}    
& \textbf{100} & \textbf{70.0} & \textbf{93.3} & \textbf{66.7}    
& \textbf{60.0} & \textbf{50.0} & \textbf{46.7} & \textbf{16.7} \\
\midrule

\multirow{5}{*}{\textbf{5}}
& GPT-4o 
& T-CoT & 66.7 & 23.3 & 56.7 & 26.7    
& 90.0 & 80.0 & 86.7 & 56.7    
& 63.3 & 40.0 & 43.3 & 20.0    
& 83.3 & 23.3 & 78.3 & 10.0    
& 40.0 & 0.0 & 33.3 & 0.0 \\
\cmidrule(lr){2-23}
& \multirow{4}{*}{Anole}
& T-CoT       & 71.7 & 16.7 & 46.7 & 16.7    
& 73.3 & 33.3 & 43.3 & 26.7    
& 60.0 & 3.3  & 50.0 & 0.0    
& 71.7 & 51.7 & 43.3 & 31.7    
& 31.7 & 0.0  & 25.0 & 0.0 \\
&
& MVoT      & 83.3 & 26.7 & 56.7 & 26.7    
& 73.3 & 43.3 & 53.3 & 36.7    
& 61.7 & 10.0  & 55.0 & 3.3    
& 73.3 & 56.7 & 53.3 & 38.3    
& 33.3 & 3.3  & 30.0 & 0.0 \\
&
& SVoT$_{\text{o}}$ & 90.0 & 76.7 & 60.0 & 33.3    
& 90.0 & 83.3 & 83.3 & 56.7    
& 70.0 & 66.7 & 63.3 & 45.0    
& 90.0 & 60.0 & 76.7 & 56.7    
& \textbf{46.7} & 0.0  & 30.0 & 0.0 \\
\rowcolor{bestgray}
&
& SVoT$_{\text{p}}$ & \textbf{93.3} & \textbf{80.0} & \textbf{70.0} & \textbf{41.7}    
& \textbf{91.7} & \textbf{86.7} & \textbf{85.0} & \textbf{58.3}    
& \textbf{78.3} & \textbf{71.7} & \textbf{73.3} & \textbf{53.3}    
& \textbf{93.3} & \textbf{66.7} & \textbf{83.3} & \textbf{60.0}    
& \textbf{46.7} & \textbf{3.3}  & \textbf{33.3} & \textbf{1.7} \\
\midrule

\multirow{5}{*}{\textbf{6}}
& GPT-4o 
& T-CoT & 63.3 & 23.3 & 40.0 & 16.7    
& 86.7 & 60.0 & 76.7 & 53.3    
& 45.0 & 36.7 & 35.0 & 16.7    
& 76.7 & 10.0 & 60.0 & 8.3    
& 36.7 & 3.3 & 26.7 & 0.0 \\
\cmidrule(lr){2-23}
& \multirow{4}{*}{Anole}
& T-CoT       & 70.0 & 16.7 & 36.7 & 1.7     
& 50.0 & 36.7 & 33.3 & 13.3    
& 58.3 & 3.3  & 41.7 & 3.3    
& 48.3 & 36.7 & 40.0 & 28.3    
& 38.3 & 0.0  & 23.3 & 0.0 \\
&
& MVoT      & 83.3 & 26.7 & 46.7 & 3.3     
& 63.3 & 40.0 & 43.3 & 20.0    
& 58.3 & 6.7  & 51.7 & 6.7    
& 60.0 & 45.0 & 51.7 & 26.7    
& 40.0 & 0.0  & 20.0 & 0.0 \\
&
& SVoT$_{\text{o}}$ & 86.7 & 70.0 & 55.0 & 26.7    
& \textbf{93.3} & 80.0 & \textbf{76.7} & 40.0    
& 66.7 & 55.0 & 56.7 & 33.3    
& \textbf{83.3} & 53.3 & 68.3 & 41.7    
& \textbf{46.7} & 3.3  & 23.3 & 0.0 \\
\rowcolor{bestgray}
&
& SVoT$_{\text{p}}$ & \textbf{90.0} & \textbf{81.7} & \textbf{66.7} & \textbf{36.7}    
& \textbf{93.3} & \textbf{83.3} & \textbf{76.7} & \textbf{55.0}    
& \textbf{71.7} & \textbf{68.3} & \textbf{65.0} & \textbf{48.3}    
& \textbf{83.3} & \textbf{60.0} & \textbf{75.0} & \textbf{56.7}    
& 45.0 & \textbf{6.7}  & \textbf{30.0} & 1.7 \\
\midrule

\multirow{5}{*}{\textbf{7}}
& GPT-4o 
& T-CoT & 66.7 & 26.7 & 36.7 & 10.0    
& \blank & \blank & \blank & \blank    
& \blank & \blank & \blank & \blank    
& \blank & \blank & \blank & \blank    
& \blank & \blank & \blank & \blank \\
\cmidrule(lr){2-23}
& \multirow{4}{*}{Anole}
& T-CoT       & 60.0 & 3.3  & 30.0 & 0.0     
& - & - & - & -    
& - & - & - & -    
& - & - & - & -    
& - & - & - & - \\
&
& MVoT      & 73.3 & 6.7  & 40.0 & 0.0     
& - & - & - & -    
& - & - & - & -    
& - & - & - & -    
& - & - & - & - \\
&
& SVoT$_{\text{o}}$ & 76.7 & 63.3 & 51.7 & 20.0    
& - & - & - & -    
& - & - & - & -    
& - & - & - & -    
& - & - & - & - \\
\rowcolor{bestgray}
&
& SVoT$_{\text{p}}$ & \textbf{80.0} & \textbf{70.0} & \textbf{65.0} & \textbf{33.3}    
& - & - & - & -    
& - & - & - & -    
& - & - & - & -    
& - & - & - & - \\
\bottomrule
\end{tabular}
\vspace{-0.2cm}
\caption{Goal coverage accuracy (\%) across five domains under ID and OOD settings, evaluated on classification (C) and free-response (F) output formats.  SVoT variants are distinguished by subscripts $\text{o}$ (ORM) and $\text{p}$ (PRM). Best results among Anole-based methods are \textbf{bolded}, and results for SVoT$_{\text{p}}$ are highlighted in gray.}


\label{tab:results_by_size}
\vspace{-0.4cm}
\end{table*}
\section{Experiment} \label{Experiment_Section}

We fine-tune Anole with LoRA~\cite{hu2022lora}, applying updates to all attention matrices and the unified output head (rank $r=8$, scaling factor $\alpha=16$, dropout $0.1$). The training pipeline consists of 5 epochs of SFT (learning rate $1 \times 10^{-4}$), followed by 5 epochs of GRPO (learning rate $5 \times 10^{-6}$). In both stages, we maintain a batch size of 8. During GRPO, we sample $G=4$ completions with a temperature of $1.0$. We implement both SFT and GRPO with the TRL library~\cite{von2020trl} and optimize using AdamW~\cite{loshchilov2017decoupled}. The entire training process is run on 4 NVIDIA A100 GPUs and takes approximately 10 days.

We begin by evaluating goal prediction accuracy on ID and OOD test sets, each containing 120 instances for every domain--size pair. Formally, given the task specification $\mathcal{X}$, the model must predict the goal state $g$ by autoregressively generating each intermediate state $z_i$ and its corresponding visualization $v_i$ in sequence, guided by the transition reasoning chain $c_i$. We report performance under two output formats: \textit{Free Response}, where the model provides a textual description of $g$, and \textit{Classification}, where it selects the correct $g$ from four candidates (see Appendix~\ref{Candidate Answer Construction} for candidate construction details). A prediction is considered correct if it exactly matches the ground truth. We adopt the state-of-the-art solver MVoT as our primary baseline. Additionally, we include two textual CoT (T-CoT) baselines: GPT-4o~\cite{openAI2024GPT-4o}, a strong non-finetuned MLLM, and Anole, which is instruction-tuned to reason step by step by explicitly describing coordinates and the environment layout in text before producing the final answer. For SVoT, we evaluate two variants: SVoT$_{\text{o}}$ (SVoT+ORM) and SVoT$_{\text{p}}$ (SVoT+PRM). For fair comparison, all baselines are provided with the same initial state description as SVoT.

\textbf{Main Results}. As shown in Table~\ref{tab:results_by_size}, free-response evaluation is markedly harder than classification, and OOD further degrades performance across all  domains, underscoring the difficulty of long-horizon state tracking.  While MVoT outperforms the Anole T-CoT baseline, both SVoT variants substantially improve over MVoT across most domain--size pairs. By predicting structured transition reasoning chains, intermediate states, and aligned visualizations, SVoT makes state tracking explicitly verifiable and enables precise reward design for GRPO.  SVoT$_{\text{p}}$ outperforms both SVoT$_{\text{o}}$ and MVoT in most settings, achieving state-of-the-art performance, particularly under the challenging OOD and free-response settings, with a largest absolute gain of 65\% over MVoT on \textsc{Sokoban} at size 4. The superior performance of SVoT$_{\text{p}}$ suggests that transition-aware rewards provide more effective guidance than outcome-only signals for complex state tracking tasks. Compared with GPT-4o, SVoT$_{\text{p}}$ generally performs better, while achieving comparable performance on \textsc{FrozenLake}. GPT-4o shows limited free-response performance in \textsc{Pacman} and \textsc{Gather}, where numerical reasoning introduces additional difficulty, and also degrades on \textsc{Maze}, where collision detection and boundary-constraint reasoning impose significant challenges. Notably, \textsc{Gather} remains challenging for all methods despite the relative advantage of SVoT$_{\text{p}}$. To diagnose this, we analyze single-step prediction accuracy to isolate individual error sources, with results discussed below.

\begin{wraptable}{r}{0.53\textwidth}
\centering
\scriptsize 
\setlength{\tabcolsep}{1.5pt}
\renewcommand{\arraystretch}{1}
\begin{tabular}{@{} l l l ccc ccc @{}}
\toprule
\textbf{Size} & \textbf{Setting} ($N$) & \textbf{Method} 
& \multicolumn{3}{c}{\makecell[b]{\textbf{Transition Reasoning}\\\textbf{Chain}}} 
& \multicolumn{3}{c}{\makecell[b]{\textbf{Intermediate}\\\textbf{State}}} \\ 
\cmidrule(lr){4-6} \cmidrule(lr){7-9}
& & & PosOld & PosNew & Ball & Action & Pos & Ball \\
\midrule

\multirow{4}{*}{4} 
& \multirow{2}{*}{ID (646)} 
&SVoT$_{\text{o}}$ &97.83 & \textbf{98.92} &72.76 &\textbf{100.0} &\textbf{97.68} &68.27 \\
& & \cellcolor{gray!15}SVoT$_{\text{p}}$ & \cellcolor{gray!15}\textbf{98.92} &\cellcolor{gray!15} 93.65 &\cellcolor{gray!15}\textbf{74.46} & \cellcolor{gray!15}\textbf{100.0} & \cellcolor{gray!15}93.34 & \cellcolor{gray!15}\textbf{74.92} \\
\cmidrule{2-9}
& \multirow{2}{*}{OOD (1196)} 
& SVoT$_{\text{o}}$ & 96.74 & \textbf{93.48} & 40.89 & 87.79 & 90.89 & 40.89 \\
& & \cellcolor{gray!15}SVoT$_{\text{p}}$ & \cellcolor{gray!15}\textbf{97.83} & \cellcolor{gray!15}92.39 & \cellcolor{gray!15}\textbf{57.36} & \cellcolor{gray!15}\textbf{95.07} & \cellcolor{gray!15}\textbf{91.39} & \cellcolor{gray!15}\textbf{51.09} \\

\midrule

\multirow{4}{*}{5} 
& \multirow{2}{*}{ID (743)} 
& SVoT$_{\text{o}}$ & \textbf{97.98} & 94.75 & 50.87 & \textbf{95.42} & 92.19 & 48.05 \\
& & \cellcolor{gray!15}SVoT$_{\text{p}}$ & \cellcolor{gray!15}\textbf{97.98} & \cellcolor{gray!15}\textbf{95.83} & \cellcolor{gray!15}\textbf{55.99} & \cellcolor{gray!15}\textbf{95.42} & \cellcolor{gray!15}\textbf{95.42} & \cellcolor{gray!15}\textbf{55.99} \\
\cmidrule{2-9}
& \multirow{2}{*}{OOD (1275)} 
& SVoT$_{\text{o}}$ & \textbf{96.78} & 93.49 & 40.47 & 93.80 & 92.55 & 40.47 \\
& & \cellcolor{gray!15}SVoT$_{\text{p}}$ &\cellcolor{gray!15}\textbf{96.78} & \cellcolor{gray!15}\textbf{94.67} & \cellcolor{gray!15}\textbf{47.84} & \cellcolor{gray!15}\textbf{95.84} & \cellcolor{gray!15}\textbf{93.80} & \cellcolor{gray!15}\textbf{44.78} \\

\midrule

\multirow{4}{*}{6} 
& \multirow{2}{*}{ID (951)} 
& SVoT$_{\text{o}}$ & 96.85 & 90.43 & 40.90 & 94.01 & 90.22 & 35.12 \\
& & \cellcolor{gray!15}SVoT$_{\text{p}}$ & \cellcolor{gray!15}\textbf{98.95} & \cellcolor{gray!15}\textbf{95.69} & \cellcolor{gray!15}\textbf{45.95} & \cellcolor{gray!15}\textbf{98.00} & \cellcolor{gray!15}\textbf{94.64} & \cellcolor{gray!15}\textbf{45.95} \\
\cmidrule{2-9}
& \multirow{2}{*}{OOD (1418)} 
& SVoT$_{\text{o}}$ & 94.64 & \textbf{94.64} & 31.73 & 88.58 & 83.29 & 31.73 \\
& & \cellcolor{gray!15}SVoT$_{\text{p}}$ &\cellcolor{gray!15}\textbf{95.70} & \cellcolor{gray!15}\textbf{94.64} & \cellcolor{gray!15}\textbf{44.71} & \cellcolor{gray!15}\textbf{92.88} & \cellcolor{gray!15}\textbf{89.07} & \cellcolor{gray!15}\textbf{41.47} \\
\bottomrule
\end{tabular}

\caption{Single-step prediction accuracy (\%) of the transition reasoning chain and intermediate state for SVoT$_{\text{o}}$ and SVoT$_{\text{p}}$ within the \textsc{Gather} domain.}
\label{tab:cot_state_results}
\end{wraptable}

\textbf{Single-Step Prediction Accuracy}. We further investigate the performance drop in \textsc{Gather} by reporting single-step prediction accuracy for both transition reasoning chain $c_i$ and intermediate state $z_i$, conditioning on ground-truth inputs. Results are measured over $N$ state predictions across 120 ID and OOD test instances for each grid size. For $c_i$, we assess the player’s position before a multi-step movement (PosOld), after the movement (PosNew), and the ball-collection reasoning (e.g., “collects 1 green ball and 1 blue ball”). For $z_i$, we evaluate the predicted action, the structured player position (Pos), and the record of collected balls (e.g., “Green: 1, Blue: 1”).

As shown in Table~\ref{tab:cot_state_results}, the main bottleneck lies in path-dependent ball collection rather than spatial tracking. SVoT maintains robust fidelity in player dynamics (action and position prediction) across all settings, with only a minor drop in position accuracy as grid size increases. 
In contrast, ball-tracking accuracy degrades sharply with increasing grid size and drops further under OOD shift for both SVoT$_{\text{o}}$ and SVoT$_{\text{p}}$. SVoT$_{\text{p}}$ maintains ball-tracking accuracy above 40\% in both $c_i$ and $z_i$ predictions. Nevertheless, multi-step collection of multi-colored balls introduces ambiguity and amplifies error accumulation over long horizons, contributing to the low performance reported in Table~\ref{tab:results_by_size}.

\setlength{\intextsep}{0pt} %
\begin{wrapfigure}{R}{0.53\textwidth}
    \centering
    \setlength{\abovecaptionskip}{0.5pt}
    \includegraphics[width=0.53\textwidth]{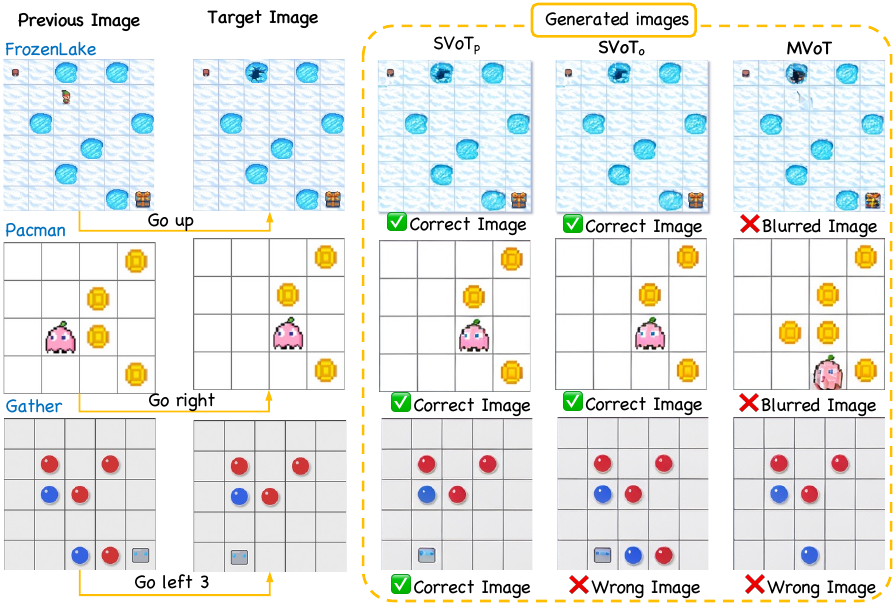}
        \caption{Example visualizations generated in Table \ref{tab:results_by_size} under the OOD and free-response settings.}
    \label{fig:images}
\end{wrapfigure}

\textbf{Visualization Qualitative Analysis}. 
Figure~\ref{fig:images} illustrates representative visualizations generated under the OOD and free-response settings in Table~\ref{tab:results_by_size}. MVoT-generated images are prone to blurring as pattern complexity increases (e.g., in \textsc{FrozenLake} and \textsc{Pacman}) and often exhibit inaccuracies (e.g., in \textsc{Gather}). In contrast, SVoT mitigates blur by penalizing vague generations via the visual reward. Notably, SVoT$_{\text{p}}$ produces high-fidelity visualizations; for instance, in \textsc{Gather}, its transition reasoning chain correctly tracks the ball-collection event (e.g., ``collects 1 red ball and 1 blue ball''), and the visualization accurately reflects this collection dynamic. Conversely, SVoT$_{\text{o}}$ suffers from an under-optimized reasoning chain (e.g., ``balls remain unchanged''), resulting in a generated image that updates the agent’s position but omits the  corresponding ball collection.

\begin{wraptable}{r}{0.52\textwidth}
    \centering
    \small 
    \setlength{\tabcolsep}{1pt} 
    \renewcommand{\arraystretch}{1} 
    
    \begin{tabular}{l !{\color{lightgray}\vrule} c !{\color{lightgray}\vrule} c !{\color{lightgray}\vrule} c !{\color{lightgray}\vrule} c !{\color{lightgray}\vrule} c}
        \toprule
        \textbf{Method} 
        & \textbf{Maze} &\textbf{FrozenLake} & \textbf{Sokoban} & \textbf{Pacman} & \textbf{Gather} \\
        \midrule
        
        MVoT 
        & 0.8056 & 0.8361 & 0.7962 & 0.7954 & 0.7414 \\
        
        SVoT$_{\text{o}}$
        & 0.8243 & 0.8464 & 0.8238 & 0.8149 & 0.7482 \\
        
        \rowcolor{gray!15}
        SVoT$_{\text{p}}$
        & \textbf{0.8571} & \textbf{0.8576} & \textbf{0.8578} & \textbf{0.8449} & \textbf{0.7716} \\
        \bottomrule
    \end{tabular}
\caption{Quantitative analysis of generated images in Table \ref{tab:results_by_size} aggregated by size under OOD and free-response settings. Scores are computed using Eq.~\eqref{eq:image_reward}. See Appendix~\ref{Visualization Quantitative Analysis Breakdown by Grid Size} for additional experiments.
}
\label{tab:main_averaged_results}
\end{wraptable}

\textbf{Visualization Quantitative Analysis}. Table~\ref{tab:main_averaged_results} reports quantitative evaluation of generated images in Table~\ref{tab:results_by_size} under OOD and free-response settings, computed using the visual reward $r_v$ (Eq.~\eqref{eq:image_reward}). Across all domains, SVoT achieves higher $r_v$ scores than MVoT, reflecting the effectiveness of $r_v$ in guiding generation. Between the two variants, SVoT$_{\text{p}}$ consistently outperforms SVoT$_{\text{o}}$ despite sharing the same visual reward. This suggests that PRM produces more faithful and generalizable visualizations by promoting accurate state-transition reasoning. In \textsc{Gather}, the lower scores of both SVoT variants align with the reduced transition-reasoning and intermediate-state prediction accuracy in Table~\ref{tab:cot_state_results}, indicating that visual quality is closely tied to the correctness of the underlying transition reasoning and state representation.

\begin{wraptable}{r}{0.57\textwidth}

\centering
\scriptsize
\setlength{\tabcolsep}{1.5pt} 
\renewcommand{\arraystretch}{1}

\newcolumntype{V}{!{\color{lightgray}\vrule}}
\begin{tabular}{@{} c l cc V cc V cc V cc V cc @{}}
\toprule
\multirow{2}{*}{\textbf{Size}} & \multirow{2}{*}{\textbf{Method}}
& \multicolumn{2}{c}{\textbf{Maze}}
& \multicolumn{2}{c}{\textbf{FrozenLake}}
& \multicolumn{2}{c}{\textbf{Sokoban}}
& \multicolumn{2}{c}{\textbf{Pacman}}
& \multicolumn{2}{c}{\textbf{Gather}} \\
\cmidrule(lr){3-4}\cmidrule(lr){5-6}\cmidrule(lr){7-8}\cmidrule(lr){9-10}\cmidrule(lr){11-12}
& 
& \textbf{ID} & \textbf{OOD}
& \textbf{ID} & \textbf{OOD}
& \textbf{ID} & \textbf{OOD}
& \textbf{ID} & \textbf{OOD}
& \textbf{ID} & \textbf{OOD} \\
\midrule

\rowcolor{gray!15}
\multirow{5}{*}{\textbf{4}}
& SVoT$_{\text{p}}$
& - & - & 83.3 & 68.3 & 78.3 & 68.3 & 70.0 & 66.7 & 50.0 & 16.7 \\
& \textit{w/o}-V
& - & - & 73.3 & 63.3 & 58.3 & 43.3 & 65.0 & 53.3 & 35.0 & 3.3 \\
& \textit{w/o}-RL
& - & - & 78.3 & 65.0 & 35.0 & 25.0 & 61.7 & 48.3 & 23.3 & 8.3 \\
& \textit{w/o}-RL-C
& - & - & 73.3 & 60.0 & 30.0 & 26.7 & 58.3 & 46.7 & 23.3 & 6.7 \\ \cmidrule(lr){2-12}
& MVoT
& - & - & 46.7 & 56.7 & 15.0 & 3.3 & 56.7 & 46.7 & 13.3 & 3.3 \\
\midrule

\rowcolor{gray!15}
\multirow{5}{*}{\textbf{5}}
& SVoT$_{\text{p}}$
& 80.0 & 41.7 & 86.7 & 58.3 & 71.7 & 53.3 & 66.7 & 60.0 & 3.3 & 1.7 \\
& \textit{w/o}-V
& 68.3 & 35.0 & 70.0 & 53.3 & 53.3 & 40.0 & 61.7 & 50.0 & 0.0 & 0.0 \\
& \textit{w/o}-RL
& 33.3 & 28.3 & 65.0 & 50.0 & 33.3 & 21.7 & 55.0 & 46.7 & 3.3 & 1.7 \\
& \textit{w/o}-RL-C
& 36.7 & 23.3 & 50.0 & 36.7 & 23.3 & 16.7 & 53.3 & 41.7 & 1.7 & 0.0 \\\cmidrule(lr){2-12}
& MVoT
& 26.7 & 26.7 & 43.3 & 36.7 & 10.0 & 3.3 & 56.7 & 38.3 & 3.3 & 0.0 \\
\midrule

\rowcolor{gray!15}
\multirow{5}{*}{\textbf{6}}
& SVoT$_{\text{p}}$
& 81.7 & 36.7 & 83.3 & 55.0 & 68.3 & 48.3 & 60.0 & 56.7 & 6.7 & 1.7 \\
& \textit{w/o}-V
& 70.0 & 25.0 & 63.3 & 45.0 & 45.0 & 26.7 & 55.0 & 45.0 & 0.0 & 0.0 \\
& \textit{w/o}-RL
& 38.3 & 16.7 & 51.7 & 38.3 & 26.7 & 16.7 & 51.7 & 40.0 & 6.7 & 1.7 \\
& \textit{w/o}-RL-C
& 33.3 & 16.7 & 40.0 & 28.3 & 20.0 & 16.7 & 50.0 & 30.0 & 3.3 & 0.0 \\
\cmidrule(lr){2-12}
& MVoT
& 26.7 & 3.3 & 40.0 & 20.0 & 6.7 & 6.7 & 45.0 & 26.7 & 0.0 & 0.0 \\
\bottomrule
\end{tabular}
\vspace{-0.2cm}
\caption{Ablation study of SVoT$_\text{p}$ evaluated on goal coverage accuracy, using the same settings as Table \ref{tab:results_by_size}. \textit{w/o}-V denotes SVoT$_\text{p}$ without visualizations; \textit{w/o}-RL denotes SVoT$_\text{p}$ without GRPO (trained using only SFT for 10 epochs); and \textit{w/o}-RL-C represents SFT with only state representations and visualizations without transition reasoning chains. See Appendix \ref{Detailed Ablation Study} for the whole table.}
\label{tab:ablation_study}
\end{wraptable}

\textbf{Ablation Study}.  Table \ref{tab:ablation_study} details the ablation study of SVoT$_{\text{p}}$ under the same experimental settings as Table \ref{tab:results_by_size}. We report free-response results for grid sizes 4 to 6 to save space. The removal of visualizations (\textit{w/o}-V) leads to a clear performance decline, suggesting that interleaved images provide essential visual cues to facilitate state tracking. Replacing GRPO with an extended SFT stage (\textit{w/o}-RL), while keeping the same training-epoch budget, results in a substantial performance drop. This is evident in the spatially constrained \textsc{Maze} domain and the interaction-rich \textsc{Sokoban} domain, where detecting invalid moves is critical and multiple player--block interactions frequently occur. These results underscore the advantage of GRPO over SFT in reasoning under complex constraints and rich interaction dynamics. Finally, after further removing transition reasoning chains from \textit{w/o}-RL, \textit{w/o}-RL-C yields a marginal decrease, indicating that SFT alone is insufficient to fully exploit transition reasoning chains without RL-based reward optimization.

\setlength{\intextsep}{0pt} %
\begin{wrapfigure}{R}{0.51\textwidth}
    \centering
        \setlength{\abovecaptionskip}{0pt}
\includegraphics[width=0.5\textwidth]{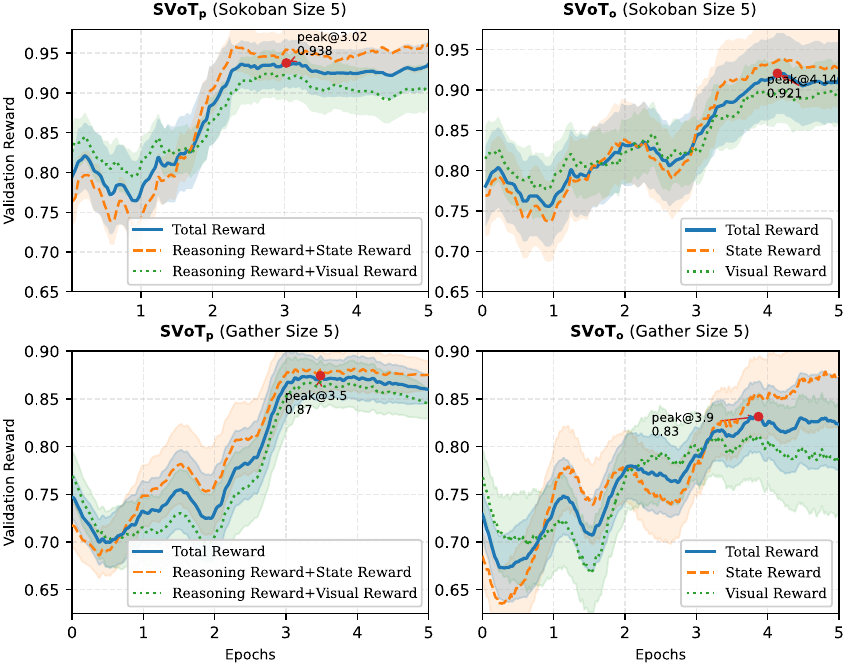}
\caption{Reward curves of
SVoT$_\text{p}$ and SVoT$_\text{o}$ on the \textsc{Sokoban} and \textsc{Gather} validation sets for single-step prediction (876 and 742 samples, respectively). Shaded regions indicate standard deviation.}

    \label{fig:reward_curve}
\end{wrapfigure}

\textbf{Reward Analysis}. Figure~\ref{fig:reward_curve} illustrates the validation reward curves of SVoT$_\text{p}$ and SVoT$_\text{o}$ for single-step prediction on \textsc{Sokoban} and \textsc{Gather} with grid size $5$, including the state reward $r_z$, visual reward $r_v$, and reasoning reward $r_c$. Under SVoT$_\text{o}$, the total reward is $0.5\,r_z + 0.5\,r_v$; under SVoT$_\text{p}$, the total reward averages $0.5\,r_c+0.5\,r_z$ and $0.5\,r_c+0.5\,r_v$. Overall, SVoT with PRM yields faster and stable optimization than ORM, due to its denser, reasoning-aware supervision. In \textsc{Sokoban}, SVoT$_\text{p}$ reaches peak total reward at around epoch 3, while SVoT$_\text{o}$ attains a comparable level  around epoch 4. For the challenging \textsc{Gather} domain, the lower peak total reward under SVoT$_\text{p}$ ($0.87$) reflects its higher task complexity. Under SVoT$_\text{o}$, the state reward $r_z$ and visual reward $r_v$ diverge post-peak, suggesting a decoupling between textual and visual representations, where the model appears to optimize textual patterns independently of visual consistency.

\section{Conclusion}

In this work, we introduced SVoT, an RL-based framework that generates interleaved, verifiable intermediate states and visualizations and treats state transitions as explicit reasoning steps to improve multi-hop spatial reasoning in MLLMs. This design bridges perception-driven spatial understanding with deterministic transition reasoning for reliable sequential state tracking. By establishing five grid-based domains that support precise verification of intermediate-state and transition-reasoning correctness, we enabled fine-grained reward optimization. Our results show that accurate intermediate-state prediction and faithful transition reasoning are central to reliable sequential state tracking, with transition-aware supervision yielding state-of-the-art performance across all evaluated domains.

\bibliographystyle{unsrtnat}
\bibliography{reference}


\clearpage
\newpage
\appendix
\section{Appendix}

\subsection{Related Work} \label{Related Work}

\textbf{Static Spatial Reasoning}. Despite significant advancements in Large Language Models (LLMs), Vision-Language Models (VLMs), and Multimodal Large Language Models
(MLLMs), spatial reasoning remains a persistent challenge and a growing area of interest \cite{liu2023visual}. Previous research attributes the limited spatial capabilities of current models to a scarcity of explicit 3D and 2D spatial knowledge in training data. To address this, recent works have introduced comprehensive data generation pipelines \cite{zhang-etal-2024-countercurate,chen2024spatialvlm,yuan2024robopoint,NEURIPS2024_f38cb4cf} and datasets \cite{wang2024picture,li2024topviewrs} to enable large-scale training on spatially-aware Visual Question Answering (VQA) tasks. Beyond data augmentation strategies, recent architectural advancements have focused on enhancing static spatial perception. SpatialRGPT \cite{NEURIPS2024_f38cb4cf} introduces a relative-depth injection module and region proposals to ground objects in 3D space, while Spatial-MLLM \cite{wu2025spatial} leverages a dual-encoder architecture to recover implicit structural priors from 2D inputs without external depth data. However, these approaches primarily target instantaneous spatial understanding, localizing objects and inferring relations within a single, static frame.  They rarely model environment dynamics during action execution, limiting their effectiveness for complex multi-hop spatial reasoning.

\textbf{Dynamic Spatial Reasoning}. To enhance dynamic spatial reasoning in LLMs, \citet{wu2024mind} proposed Visualization-of-Thought (VoT), a zero-shot prompting strategy that encourages models to internally simulate spatial layouts and their evolution over time. Inspired by human mental navigation, VoT prompts the model to generate a text-based state description after each reasoning step to guide multi-step inference. Recent strides in multimodal-native foundation models, such as Chameleon \cite{team2024chameleon}, Emu \cite{cui2025emu3}, and Show-o \cite{xie2025show}, have expanded capabilities from passive understanding to active multimodal generation. Building on the unified multimodal backbone, Multimodal Visualization-of-Thought (MVoT)~\cite{li2025imagine} advances VoT by moving beyond purely textual reasoning. By synthesizing visual tokens directly within the reasoning chain, MVoT integrates text and vision into an interleaved reasoning format. This provides a visually grounded reasoning trace, making the intermediate decision-making process more explicit. However, a key limitation of both VoT and MVoT is the absence of a principled mechanism to verify intermediate steps. As a result, errors can accumulate over time, undermining reliable sequential state tracking in multi-hop spatial reasoning. In contrast, SVoT introduces verifiable structured state representations and explicit transition reasoning chains, allowing intermediate states and reasoning chains to be optimized via reinforcement learning. By tightly coupling transition reasoning, state prediction, and visualization generation, SVoT mitigates error propagation and improves the consistency of long-horizon state tracking.

\textbf{Reinforcement Learning for Reasoning}. Reinforcement learning (RL) has emerged as an effective approach for enhancing reasoning in large models, as demonstrated by OpenAI o1~\cite{jaech2024openai}, DeepSeek-R1~\cite{guo2025deepseek}, and Qwen3~\cite{yang2025qwen3}. Among recent algorithms, Group Relative Policy Optimization (GRPO)~\cite{shao2024deepseekmath} provides a computationally efficient approach by normalizing rewards over a group of sampled candidate outputs, yielding stable learning signals without requiring a separate critic model. This paradigm has recently been extended to multimodal applications~\cite{jiang2025propa, deng2025openvlthinker, xu2025visual, zhang2025r1}. Building on these advances, SVoT adapts GRPO to improve sequential state tracking in multi-hop spatial reasoning by optimizing an interleaved generation process that produces structured intermediate states and visualizations guided by transition reasoning chains. With fine-grained rewards, SVoT verifies predicted intermediate states and transition reasoning chains while evaluating visualization fidelity, achieving state-of-the-art performance on challenging sequential state tracking tasks with strong out-of-distribution (OOD) robustness.

\subsection{MVoT} \label{MVoT}
The supervised fine-tuning (SFT) stage of SVoT follows the same training pipeline as MVoT. MVoT builds on Chameleon’s unified autoregressive framework~\cite{team2024chameleon}, where image and text tokens are jointly processed by a single causal Transformer. Images are encoded by a frozen visual tokenizer with codebook $C \in \mathbb{R}^{N \times D}$, where $N$ denotes the number of codebook entries and $D$ denotes the embedding dimensionality. We denote the $i$-th visual codebook index as $t_i^{\mathrm{vis}} \in \{1,\dots,N\}$ and the embedding associated with index $j$ as $e_j^{\mathrm{vis}} \in \mathbb{R}^D$. Text is mapped to discrete token indices by a separate frozen text tokenizer, and the resulting text and image token sequences are concatenated before being fed into the Transformer.

In addition to the standard next-token cross-entropy objective, MVoT incorporates a \emph{token discrepancy loss} that explicitly regularizes the model in the visual embedding space.
For each ground-truth visual index $t_i^{\mathrm{vis}}$, MVoT first computes the distance between its corresponding embedding $e^{\mathrm{vis}}_{t_i^{\mathrm{vis}}}$ and all visual codebook entries to obtain
\begin{equation}
    S_{t_i^{\mathrm{vis}}}
    =
    \bigl[
        \mathrm{MSE}(e^{\mathrm{vis}}_{t_i^{\mathrm{vis}}}, e^{\mathrm{vis}}_1),
        \dots,
        \mathrm{MSE}(e^{\mathrm{vis}}_{t_i^{\mathrm{vis}}}, e^{\mathrm{vis}}_N)
    \bigr]
    \in \mathbb{R}^{1 \times N},
    \label{eq:svot-token-sim}
\end{equation}
where $\mathrm{MSE}(\cdot,\cdot)$ denotes the mean squared error and larger values indicate that a codebook entry is farther from the ground-truth visual embedding.
At position $i$, the Transformer outputs a probability distribution $P(t_i) \in \mathbb{R}^{1 \times N}$ over the visual vocabulary.
The token discrepancy loss is then defined as
\begin{equation}
    \mathcal{L}_D = \sum_{i=1}^{N} S_{t_i^{\mathrm{vis}}} \cdot P(t_i),
    \label{eq:svot-discrepancy-loss}
\end{equation}
where $\cdot$ denotes the inner product.
Intuitively, $\mathcal{L}_D$ discourages the model from assigning high probability to codebook entries that are associated with large distances to the ground-truth embedding in the visual space.

The causal Transformer is fine-tuned with a combination of the standard next-token cross-entropy loss $\mathcal{L}_C$ over both text and image tokens and the token discrepancy loss $\mathcal{L}_D$ on image tokens, while the image and text tokenizers remain frozen. The overall training objective is
\begin{equation}
\mathcal{L} = \mathcal{L}_C + \mathcal{L}_D.
\end{equation}

\subsection{Detailed Formulation of Visual Reward} \label{Detailed Formulation of Visual Reward}
\label{app:visual_metrics}

The visual reward $r_v$ is defined as
{
\setlength{\abovedisplayskip}{6pt}
\setlength{\belowdisplayskip}{6pt}
\begin{align}
    r_v(\hat{v}, v) &= r_{\mathrm{str}}(\hat{v}, v)\cdot r_{\mathrm{qua}}(\hat{v}, v), \\
    r_{\mathrm{str}}(\hat{v}, v) &= \sum_{(i,j)} w_{ij}\,\mathbb{I}\left[\mathcal{S}(\hat{v}_{ij}, v_{ij}) \ge \tau \right], \\
    r_{\mathrm{qua}}(\hat{v}, v) &= \mathrm{clip}\left(\frac{\phi(\hat{v};\Omega)}{\phi(v;\Omega)+\epsilon}, 0, 1\right).
\end{align}
}

The term $r_{\mathrm{str}}$ evaluates cell-level matching, while $r_{\mathrm{qua}}$ measures perceptual quality through foreground sharpness and penalizes blurry predictions. Foreground cells are identified using a variance-based heuristic. For each cell, RGB values are normalized to the range $[0,1]$, and the cell is classified as foreground if the average per-channel variance of the normalized RGB values exceeds a threshold $\delta=0.002$. The foreground mask $\Omega$ is extracted from the ground-truth image $v$ and kept fixed when evaluating both $\hat{v}$ and $v$. It is used both to assign higher weights to semantically salient cells in $r_{\mathrm{str}}$ and to define the sharpness region in $r_{\mathrm{qua}}$.

In the structural term $r_{\mathrm{str}}$, a cell is counted as matched only when its composite similarity $\mathcal{S}$ exceeds the threshold $\tau = 0.8$, making the matching criterion relatively strict and reducing reward for coarse but inaccurate alignments. Before normalization, foreground cells are assigned weights ten times larger than those of background cells, i.e.,
\begin{equation}
w_{ij} \propto
\begin{cases}
10, & (i,j)\in\Omega,\\
1, & (i,j)\notin\Omega,
\end{cases}
\qquad \text{with} \qquad \sum_{i,j} w_{ij}=1.
\end{equation}

We next detail the calculation of the composite similarity score $\mathcal{S}(\hat{v}_{ij}, v_{ij})$. For notational brevity, let $\mathbf{x}, \mathbf{y} \in \mathbb{R}^{H_c \times W_c \times 3}$ denote the pixel arrays of the generated cell $\hat{v}_{ij}$ and the ground-truth cell $v_{ij}$, respectively, where $H_c$ and $W_c$ are the height and width of each cell patch. $\mathcal{S}$ comprises three components:

\textbf{Perceptual Hash Similarity ($s_{\mathrm{hash}}$)}.
This component captures the coarse geometric structure of the image patch. We first convert the patches $\mathbf{x}$ and $\mathbf{y}$ to grayscale and downsample them to a fixed size of $K \times K$ (with $K=8$). Let $\mathbf{H}(\mathbf{x}) \in \{0, 1\}^{K^2}$ be the binary hash vector, where the $k$-th bit is set to 1 if the pixel intensity exceeds the patch's mean intensity, and 0 otherwise. The similarity is calculated using the normalized Hamming distance:
\begin{equation}
    s_{\mathrm{hash}}(\mathbf{x}, \mathbf{y}) = 1 - \frac{1}{K^2} \sum_{k=1}^{K^2} \mathbb{I}\left[ \mathbf{H}(\mathbf{x})_k \neq \mathbf{H}(\mathbf{y})_k \right],
\end{equation}
where $\mathbb{I}[\cdot]$ is the indicator function.

\textbf{Edge Alignment ($s_{\mathrm{edge}}$)}.
To ensure the alignment of local boundaries and textures, we extract edge maps using the Sobel operator. Let $G_x$ and $G_y$ be the horizontal and vertical gradients of the grayscale patch, computed via convolution. We define a binary edge map $\mathbf{E}(\mathbf{x})$ by thresholding the gradient magnitude:
\begin{equation}
    \mathbf{E}(\mathbf{x}) = \mathbb{I}\left[ \sqrt{G_x^2 + G_y^2} > \mu_{\mathbf{x}} \right],
\end{equation}
where $\mu_{\mathbf{x}}$ is the mean gradient magnitude of the patch. The similarity $s_{\mathrm{edge}}$ is computed as the Intersection over Union (IoU) of the binary edge maps:

\begin{equation}
    s_{\mathrm{edge}}(\mathbf{x}, \mathbf{y}) =
\begin{cases}
1, & \text{if } |\mathbf{E}(\mathbf{x}) \cup \mathbf{E}(\mathbf{y})| = 0,\\[3pt]
\dfrac{|\mathbf{E}(\mathbf{x}) \cap \mathbf{E}(\mathbf{y})|}
{|\mathbf{E}(\mathbf{x}) \cup \mathbf{E}(\mathbf{y})| + \epsilon},
& \text{otherwise}.
\end{cases}
\end{equation}
where $\epsilon$ is a small constant for numerical stability. When both edge maps are empty, we set $s_{\mathrm{edge}}=1$, treating the shared absence of edges as perfect edge-level agreement.

\textbf{Color Consistency ($s_{\mathrm{color}}$)}.
To capture chromatic correlations, we compute a joint RGB histogram. We discretize the color space into $8 \times 8 \times 8$ bins and flatten it into a vector $\mathbf{h} \in \mathbb{R}^{512}$. Let $\tilde{\mathbf{h}} = \mathbf{h} - \bar{h}\mathbf{1}$ denote the mean-centered histogram vector, where $\bar{h}$ is the mean histogram value and $\mathbf{1}$ is the all-ones vector. The similarity is measured using the Pearson correlation coefficient, formulated as the cosine similarity of the centered vectors:
\begin{equation}
    s_{\mathrm{color}}(\mathbf{x}, \mathbf{y}) = \mathrm{clip}\left( \frac{ \tilde{\mathbf{h}}_{\mathbf{x}}^\top \tilde{\mathbf{h}}_{\mathbf{y}} }{ \| \tilde{\mathbf{h}}_{\mathbf{x}} \|_2 \cdot \| \tilde{\mathbf{h}}_{\mathbf{y}} \|_2 }, 0, 1 \right).
\end{equation}

The composite similarity score $\mathcal{S}$ is computed as a weighted combination of the $s_{\mathrm{hash}}$, $s_{\mathrm{edge}}$, and $s_{\mathrm{color}}$, with $\lambda_h = 0.4$, $\lambda_e = 0.3$, and $\lambda_c = 0.3$, respectively:
\begin{equation}
\mathcal{S}(\mathbf{x}, \mathbf{y}) =
\lambda_h s_{\mathrm{hash}}(\mathbf{x}, \mathbf{y})
+ \lambda_e s_{\mathrm{edge}}(\mathbf{x}, \mathbf{y})
+ \lambda_c s_{\mathrm{color}}(\mathbf{x}, \mathbf{y}).
\end{equation}

\subsection{Group Relative Policy Optimization (GRPO)} \label{GRPO}

GRPO is a PPO-style policy optimization method that avoids training a separate value critic, and instead estimates advantages by normalizing rewards within a group of sampled completions. For each prompt $q$, GRPO samples a set of $G$ candidate completions $\{o_g\}_{g=1}^{G}$ from the current (old) policy $\pi_{\theta_{\text{old}}}(\cdot \mid q)$ and evaluates each candidate with a reward function to obtain $\{r_g\}_{g=1}^{G}$. 
The advantage of a candidate $o_g$ is computed via within group normalization:
{
\setlength{\abovedisplayskip}{2pt}
\setlength{\belowdisplayskip}{2pt}
\begin{align}
A_g = \frac{r_g - \mathrm{mean}(\{r_j\}_{j=1}^{G})}{\mathrm{std}(\{r_j\}_{j=1}^{G})}.
\label{eq:grpo_adv}
\end{align}
}

The policy $\pi_\theta$ is then updated by optimizing the following objective:

\begin{equation}
\begin{split}
\mathcal{J}_{\mathrm{GRPO}}(\theta)
&= \mathbb{E}_{q \sim \mathcal{D},\,\{o_g\}_{g=1}^{G}\sim \pi_{\theta_{\mathrm{old}}}(\cdot \mid q)} \\
&\quad \Bigg[
\frac{1}{G}\sum_{g=1}^{G}
\frac{1}{|o_g|}\sum_{t=1}^{|o_g|}
\min\!\Big(
\rho_{g,t}(\theta) A_g,\,
\mathrm{clip}(\rho_{g,t}(\theta),1-\epsilon,1+\epsilon)A_g
\Big)
-\beta\, D_{\mathrm{KL}}\!\big(\pi_{\theta}\,\|\,\pi_{\mathrm{ref}}\big)
\Bigg].
\end{split}
\label{eq:grpo_obj}
\end{equation}

where $\rho_{g,t}(\theta)=\frac{\pi_{\theta}(o_{g,t}\mid q,o_{g,<t})}{\pi_{\theta_{\mathrm{old}}}(o_{g,t}\mid q,o_{g,<t})}$ is the token-level probability ratio, $o_{g,t}$ denotes the $t$-th token of completion $o_g$, and $|o_g|$ denotes the completion length. $\epsilon$ is the clipping parameter, and $\beta$ controls the KL penalty. By normalizing rewards within each group, GRPO derives relative advantages that provide stable training signals and encourage the policy to assign higher probability to higher-advantage completions. In our experiments, we set $\epsilon=0.2$ and $\beta=0$ following prior studies \cite{hu2025open,yu2025dapo}.

\subsection{Domain Generation} \label{Domain Generation}

In this work, all domains are instantiated as OpenAI Gym environments~\cite{brockman2016openai}.
\textsc{FrozenLake} is based on the built-in Gym implementation, while the other four domains are implemented using PDDLGym~\cite{silver2020pddlgym}.
PDDLGym is a Python library that automatically builds OpenAI Gym environments from PDDL~\cite{haslum2019introduction} domain and problem files. This provides a convenient and scalable mechanism for constructing relational benchmarks for both reinforcement learning and planning-based sequential decision making. In PDDL, the domain file specifies the lifted transition model through predicate symbols and parameterized action schemas, each defined by precondition and effect formulas. The problem file specifies the concrete objects used to instantiate lifted predicates into ground atoms and action schemas into ground actions, together with the initial state, represented as a set of ground atoms, and the goal condition. When constructing datasets in our experiments, we manually design a PDDL domain file for each domain and automatically generate multiple problem files, which are used to construct the training, validation, and test sets. 

For the \textsc{Maze} domain, we generate PDDL problem files by sampling the initial player position, goal position, and wall positions under different random seeds. We discard problem files when breadth-first search (BFS) finds no feasible path from the initial player position to the goal. For each solvable instance, we construct action sequences that include both a ground-truth, goal-achieving plan obtained via BFS for the training set and randomly sampled action sequences, which may contain inapplicable moves corresponding to wall collisions or boundary violations. The wall density for each grid size is summarized in Table~\ref{tab:task_shiftmap}.

For the \textsc{FrozenLake} domain, we use the built-in OpenAI Gym generator to sample distinct maps by varying the hole density as specified in Table~\ref{tab:task_shiftmap}.
We then perform a BFS-based reachability check to discard any map whose goal state is not reachable from the initial player position. For each solvable map, we extract a  goal-reaching plan via BFS for the training set and generate additional action sequences by sampling random actions under deterministic grid dynamics (i.e., ignoring the stochastic “slippery” transitions); these trajectories may terminate early when the agent falls into a hole.
The hole density for each grid size is reported in Table~\ref{tab:task_shiftmap}.

For the \textsc{Sokoban} domain, we generate PDDL problem files by sampling the initial player position, the number of blocks, the initial and goal positions of each block, and the wall positions under different random seeds and grid sizes. Similar to the previous domains, we discard any problem file that is unsolvable and, for each solvable instance, extract a goal-achieving plan to use as supervised action sequences for training, while additionally generating random action sequences. The configuration of the number of blocks and the wall density for each grid size is reported in Table~\ref{tab:task_shiftmap}.

For the \textsc{Pacman} domain, we generate PDDL problem files by sampling the initial player position, number of coins and their locations, controlled by the coin-density configuration. Similar to the previous domains, we discard any problem file that is unsolvable and, for each solvable instance, extract a goal-achieving plan (collecting all coins) for training, while additionally generating random action sequences. The coin density for each grid size is reported in Table~\ref{tab:task_shiftmap}.

For the \textsc{Gather} domain, we generate PDDL problem files by sampling the initial player position, the number of balls of each color, and the positions of each ball, controlled by the ball-density configuration across different random seeds and grid sizes. We construct action sequences in the same manner as above, with the goal-achieving plan defined as the action sequence that collects all colored balls. The ball density and the color configuration for each grid size are reported in Table~\ref{tab:task_shiftmap}.

We present the domain descriptions of \textsc{Maze} (Table~\ref{apptab:svot-maze}),
\textsc{FrozenLake} (Table~\ref{apptab:svot-frozenlake}),
\textsc{Sokoban} (Table~\ref{apptab:svot-sokoban}),
\textsc{Pacman} (Table~\ref{apptab:svot-pacman}),
and \textsc{Gather} (Table~\ref{apptab:svot-gather})
used in the input prompts of SVoT with PRM, including the specified action {preconditions} and {effects}. We shorten the length of the action sequence to two for clarity of illustration. For \textsc{Maze}, we include both free-response and classification examples, covering three types of responses, intermediate state prediction, image prediction, and goal prediction, to illustrate the full SVoT process. For the other domains, we only show free-response goal prediction examples.

\subsection{Dataset Collection} \label{Dataset Collection}

For each domain and grid size, we construct the training set from 100 valid maps with 5 valid action sequences per map, and the validation set from 30 valid maps with 2 valid action sequences per map. We further generate separate ID and OOD test sets, each from 60 valid maps with 2 valid action sequences per map. This yields 500 unique training instances, 60 unique validation instances, 120 unique ID test instances, and 120 unique OOD test instances per domain--size pair, with no duplication within or across splits. For the OOD settings, we extend the action-sequence length and introduce additional interactive objects within each grid size to evaluate SVoT's generalization capability. Table~\ref{tab:task_shiftmap} reports training, validation, ID test, and OOD test sizes measured by the total number of state, visualization, and goal predictions, together with action-sequence length ranges and domain statistics for each grid size.

\subsection{Candidate Answer Construction} \label{Candidate Answer Construction}

For each instance in the introduced domain, we construct a four-choice classification set by pairing the ground-truth answer with three randomly generated incorrect candidates. We additionally impose constraints to ensure that one incorrect candidate is partially correct. For example, in \textsc{Pacman}, we include a distractor that either has the correct final player position but an incorrect number of collected coins, or has an incorrect final position but a correct coin count. We also constrain candidate coordinates to lie within a distance of 2 cells from the ground-truth coordinate, and require all candidates to be valid (e.g., the predicted coin count must be between 0 and the total number of coins).

\begin{figure*}
    \centering
    \includegraphics[width=1\linewidth]{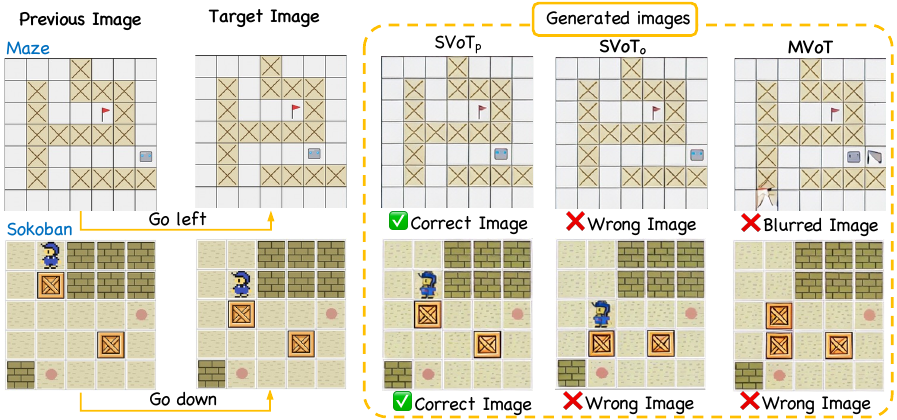}
            \vspace{-0.6cm}
        \caption{Visualization examples generated in Table \ref{tab:results_by_size} for \textsc{Maze} and \textsc{Sokoban} domains under the OOD and free-response settings.}
    
    \label{fig:images_app}
    
\end{figure*}

\begin{table}[t]
\centering
\scriptsize
\setlength{\tabcolsep}{2.4pt} 
\renewcommand{\arraystretch}{1.2} 

\begin{tabular}{@{} l l c c c c c c l l @{}} 
\toprule
\multirow{2}{*}{\textbf{Category}} 
& \multirow{2}{*}{\textbf{Domain}} 
& \multirow{2}{*}{\textbf{Grid Size}} 
& \multirow{2}{*}{\textbf{Training Size}} 
& \multirow{2}{*}{\textbf{Validation Size}} 
& \multicolumn{2}{c}{\textbf{Test Size}} 
& \multirow{2}{*}{\textbf{Color}} 
& \multirow{2}{*}{\textbf{Variables}} 
& \multirow{2}{*}{\textbf{ID $\to$ OOD}} \\
\cmidrule(lr){6-7}
& & & & & \textbf{ID} & \textbf{OOD} & & & \\
\midrule

\multirow{15}{*}{\textbf{Statics}} 
 & \multirow{3}{*}{\textsc{Maze}} 
   & 5 & 6342 & 832  & 1658 & 2682 & - 
   & \multirow{3}{*}{Wall Density} 
   & \multirow{3}{*}{$0.3 \to 0.5$} \\
 & & 6 & 8164 & 982  & 1956 & 3164 & - & & \\
 & & 7 & 8544 & 1166 & 2326 & 3372 & - & & \\
\cmidrule{2-10} 

 & \multirow{3}{*}{\textsc{FrozenLake}} 
   & 4 & 5120 & 736 & 1494 & 2362 & - 
   & \multirow{3}{*}{Hole Density} 
   & \multirow{3}{*}{$0.2 \to 0.4$} \\
 & & 5 & 7346 & 882 & 1758 & 2588 & - & & \\
 & & 6 & 7754 & 1042 & 2028 & 3018 & - & & \\
\cmidrule{2-10}

 & \multirow{3}{*}{\textsc{Sokoban}} 
   & 4 & 5326 & 762 & 1534 & 2430 & - 
   & \begin{tabular}[c]{@{}l@{}}Wall Density \\ Block Number\end{tabular} 
   & \begin{tabular}[c]{@{}l@{}}$0.2 \to 0.3$ \\ $1\text{--}2 \to 2$\end{tabular} \\
\cmidrule{3-10}
 & & 5 & 7146 & 936  & 1864 & 2704 & \multirow{2}{*}{-} 
   & \multirow{2}{*}{\begin{tabular}[c]{@{}l@{}}Wall Density \\ Block Number\end{tabular}} 
   & \multirow{2}{*}{\begin{tabular}[c]{@{}l@{}}$0.2 \to 0.3$ \\ $1\text{--}2 \to 3$\end{tabular}} \\
 & & 6 & 8362 & 1112 & 2216 & 3136 & & & \\
\cmidrule{2-10}

 & \multirow{3}{*}{\textsc{Pacman}} 
   & 4 & 5342 & 652 & 1308 & 2396 & - 
   & \multirow{3}{*}{Coin Density} 
   & \multirow{3}{*}{$0.3 \to 0.5$} \\
 & & 5 & 7426 & 846 & 1698 & 2642 & - & & \\
 & & 6 & 7728 & 934 & 1910 & 2974 & - & & \\
\cmidrule{2-10}

 & \multirow{3}{*}{\textsc{Gather}} 
   & 4 & 5424 & 696 & 1412 & 2512 & \{R, G\} 
   & \multirow{3}{*}{Ball Density} 
   & \multirow{3}{*}{$0.3 \to 0.5$} \\
\cmidrule{3-8}
 & & 5 & 7330 & 802 & 1606 & 2670 & \{R, G, B\} & & \\
\cmidrule{3-8}
 & & 6 & 8182 & 1026 & 2022 & 2956 & \{R, G, B, Y\} & & \\

\midrule
\midrule

\multirow{4}{*}{\textbf{Action}} 
 & \multirow{4}{*}{All} 
   & 4 & -- & -- & -- & -- & - 
   & Sequence Length 
   & $3\text{--}7 \to 8\text{--}10$ \\
 & & 5 & -- & -- & -- & -- & - 
   & Sequence Length 
   & $4\text{--}8 \to 9\text{--}11$ \\
 & & 6 & -- & -- & -- & -- & - 
   & Sequence Length 
   & $5\text{--}10 \to 11\text{--}13$ \\ 
 & & 7 & -- & -- & -- & -- & - 
   & Sequence Length 
   & $6\text{--}11 \to 12\text{--}14$ \\ 

\bottomrule
\end{tabular}
\vspace{0.2cm}
\caption{Domain statistics and ID$\to$OOD parameter shifts. Training, validation, ID test, and OOD test sizes are measured by the total number of state, visualization, and goal predictions from 500 training instances, 60 validation instances, 120 ID test instances, and 120 OOD test instances for each domain--size pair, respectively. Each action step contributes one state prediction and one visualization prediction, and each instance contributes one final goal prediction.}

\label{tab:task_shiftmap}
\end{table}


\begin{table*}[h]
    \centering
    \small
    \setlength{\tabcolsep}{5pt}
    \renewcommand{\arraystretch}{1.25} 
    \resizebox{\textwidth}{!}{
    \begin{tabular}{l !{\color{lightgray}\vrule} ccc !{\color{lightgray}\vrule} ccc !{\color{lightgray}\vrule} ccc !{\color{lightgray}\vrule} ccc !{\color{lightgray}\vrule} ccc}
        \toprule
        \multirow{2}{*}{Method} 
        & \multicolumn{3}{c}{Maze} 
        & \multicolumn{3}{c}{FrozenLake} 
        & \multicolumn{3}{c}{Sokoban} 
        & \multicolumn{3}{c}{Pacman} 
        & \multicolumn{3}{c}{Gather} \\
        \cmidrule(lr){2-4}\cmidrule(lr){5-7}\cmidrule(lr){8-10}\cmidrule(lr){11-13}\cmidrule(lr){14-16}
        & 5 & 6 & 7 & 4 & 5 & 6 & 4 & 5 & 6 & 4 & 5 & 6 & 4 & 5 & 6 \\
        \midrule
        
        MVoT 
        & 0.8111 & 0.8030 & 0.8028
        & 0.8329 & 0.8475 & 0.8278
        & 0.8064 & 0.8075 & 0.7748
        & 0.8064 & 0.8033 & 0.7766
        & 0.7646 & 0.7590 & 0.7005 \\
        
        SVoT$_\text{o}$
        & 0.8385 & 0.8265 & 0.8080
        & \textbf{0.8594} & 0.8393 & 0.8406
        & 0.8397 & 0.8208 & 0.8109
        & 0.8219 & 0.8217 & 0.8011
        & 0.7713 & 0.7687 & \textbf{0.7047} \\

        \rowcolor{gray!15}
        SVoT$_\text{p}$
        & \textbf{0.8748} & \textbf{0.8610} & \textbf{0.8356}
        & 0.8560 & \textbf{0.8679} & \textbf{0.8490}
        & \textbf{0.8486} & \textbf{0.8413} & \textbf{0.8834}
        & \textbf{0.8397} & \textbf{0.8444} & \textbf{0.8506}
        & \textbf{0.8190} & \textbf{0.7939} & 0.7020 \\
        \bottomrule
    \end{tabular}
    }
    \caption{Full quantitative analysis of generated images in Table \ref{tab:results_by_size} under OOD and free-response settings. Scores are  computed using the visual reward $r_v$ defined in Eq.~\eqref{eq:image_reward}}
        
    \label{tab:appendix_full_results_styled}
\end{table*}

\begin{table*}[h]
    \centering
    \small
    \setlength{\tabcolsep}{4pt}
    \renewcommand{\arraystretch}{1.18}
    \resizebox{\textwidth}{!}{
    \begin{tabular}{ll !{\color{lightgray}\vrule} ccc !{\color{lightgray}\vrule} ccc !{\color{lightgray}\vrule} ccc !{\color{lightgray}\vrule} ccc !{\color{lightgray}\vrule} ccc}
        \toprule
        \multirow{2}{*}{Method}
        & \multirow{2}{*}{Metric}
        & \multicolumn{3}{c}{Maze}
        & \multicolumn{3}{c}{FrozenLake}
        & \multicolumn{3}{c}{Sokoban}
        & \multicolumn{3}{c}{Pacman}
        & \multicolumn{3}{c}{Gather} \\
        \cmidrule(lr){3-5}
        \cmidrule(lr){6-8}
        \cmidrule(lr){9-11}
        \cmidrule(lr){12-14}
        \cmidrule(lr){15-17}
        & & 5 & 6 & 7 & 4 & 5 & 6 & 4 & 5 & 6 & 4 & 5 & 6 & 4 & 5 & 6 \\
        \midrule

        \multirow{3}{*}{MVoT}
        & FGAcc
        & 87.10 & 85.90 & 84.80
        & 88.40 & 89.20 & 87.60
        & 85.60 & 84.90 & 82.10
        & 86.40 & 85.80 & 83.10
        & 80.90 & 79.60 & 73.90 \\
        & BkgErr
        & 6.60 & 7.10 & 7.70
        & 5.90 & 5.40 & 6.20
        & 7.80 & 8.20 & 9.50
        & 7.30 & 7.70 & 9.00
        & 10.70 & 11.30 & 14.10 \\
        & FullAcc
        & 66.70 & 62.50 & 58.30
        & 68.30 & 70.00 & 65.80
        & 60.80 & 58.30 & 53.30
        & 62.50 & 60.80 & 55.00
        & 46.70 & 43.30 & 32.50 \\
        \midrule

        \multirow{3}{*}{SVoT$_\text{o}$}
        & FGAcc
        & 90.30 & 88.80 & 86.20
        & 91.40 & 89.50 & 89.30
        & 89.60 & 87.30 & 86.10
        & 88.40 & 87.70 & 85.80
        & 82.80 & 81.50 & 74.80 \\
        & BkgErr
        & 4.90 & 5.40 & 6.60
        & 4.20 & 5.10 & 5.20
        & 5.60 & 6.80 & 7.30
        & 5.90 & 6.20 & 7.50
        & 9.10 & 9.80 & \textbf{13.20} \\
        & FullAcc
        & 75.00 & 70.80 & 64.20
        & 78.30 & 73.30 & 72.50
        & 70.00 & 65.00 & 62.50
        & 69.20 & 67.50 & 62.50
        & 54.20 & 50.00 & 36.70 \\
        \midrule

        \rowcolor{gray!15}
        & FGAcc
        & \textbf{94.10} & \textbf{93.20} & \textbf{90.60}
        & \textbf{92.00} & \textbf{92.50} & \textbf{91.60}
        & \textbf{91.80} & \textbf{91.10} & \textbf{93.20}
        & \textbf{90.40} & \textbf{90.80} & \textbf{91.10}
        & \textbf{86.80} & \textbf{84.30} & \textbf{76.10} \\

        \rowcolor{gray!15}
        & BkgErr
        & \textbf{2.90} & \textbf{3.30} & \textbf{4.50}
        & \textbf{4.00} & \textbf{3.60} & \textbf{4.30}
        & \textbf{4.30} & \textbf{4.80} & \textbf{3.40}
        & \textbf{4.80} & \textbf{4.70} & \textbf{4.30}
        & \textbf{6.70} & \textbf{8.00} & 13.60 \\

        \rowcolor{gray!15}
        \multirow{-3}{*}{SVoT$_\text{p}$}
        & FullAcc
        & \textbf{84.20} & \textbf{81.70} & \textbf{75.80}
        & \textbf{80.00} & \textbf{81.70} & \textbf{78.30}
        & \textbf{77.50} & \textbf{75.80} & \textbf{83.30}
        & \textbf{74.20} & \textbf{75.00} & \textbf{76.70}
        & \textbf{63.30} & \textbf{57.50} & \textbf{38.30} \\
        \bottomrule
    \end{tabular}
    }
    \caption{
    Foreground--background correctness evaluation of generated visualizations in Table~\ref{tab:results_by_size} under OOD and free-response settings.
    FGAcc denotes Foreground Cell Accuracy, BkgErr denotes Background Error Rate, and FullAcc denotes Full Visualization Accuracy.
    Higher values are better for FGAcc and FullAcc, while lower values are better for BkgErr.
    All results are reported as percentages.
    }
    \label{tab:fg_bg_full_visualization_metrics}
\end{table*}

\definecolor{linegray}{gray}{0.65}

\begin{table*}[t]
\centering
\scriptsize
\setlength{\tabcolsep}{3.5pt} 
\renewcommand{\arraystretch}{1.15}

\resizebox{\textwidth}{!}{
\begin{tabular}{@{} c l
    *{4}{c} !{\color{linegray}\vrule} 
    *{4}{c} !{\color{linegray}\vrule} 
    *{4}{c} !{\color{linegray}\vrule} 
    *{4}{c} !{\color{linegray}\vrule} 
    *{4}{c}                           
@{}}
\toprule
\multirow{3}{*}{\textbf{Size}} & \multirow{3}{*}{\textbf{Method}}
& \multicolumn{4}{c}{\textbf{Maze}}
& \multicolumn{4}{c}{\textbf{FrozenLake}}
& \multicolumn{4}{c}{\textbf{Sokoban}}
& \multicolumn{4}{c}{\textbf{Pacman}}
& \multicolumn{4}{c}{\textbf{Gather}} \\
\cmidrule(lr){3-6}\cmidrule(lr){7-10}\cmidrule(lr){11-14}\cmidrule(lr){15-18}\cmidrule(lr){19-22}
&
& \multicolumn{2}{c}{ID} & \multicolumn{2}{c}{OOD}
& \multicolumn{2}{c}{ID} & \multicolumn{2}{c}{OOD}
& \multicolumn{2}{c}{ID} & \multicolumn{2}{c}{OOD}
& \multicolumn{2}{c}{ID} & \multicolumn{2}{c}{OOD}
& \multicolumn{2}{c}{ID} & \multicolumn{2}{c}{OOD} \\
\cmidrule(lr){3-4}\cmidrule(lr){5-6}
\cmidrule(lr){7-8}\cmidrule(lr){9-10}
\cmidrule(lr){11-12}\cmidrule(lr){13-14}
\cmidrule(lr){15-16}\cmidrule(lr){17-18}
\cmidrule(lr){19-20}\cmidrule(lr){21-22}
&
& C & F & C & F
& C & F & C & F
& C & F & C & F
& C & F & C & F
& C & F & C & F \\
\midrule

\rowcolor{gray!15}
\cellcolor{white}\multirow{6}{*}{\textbf{4}}
& SVoT$_{\text{p}}$
& - & - & - & -
& 100.0 & 83.3 & 88.3 & 68.3
& 86.7 & 78.3 & 80.0 & 68.3
& 100.0 & 70.0 & 93.3 & 66.7
& 60.0 & 50.0 & 46.7 & 16.7 \\
& \textit{w/o}-V
& - & - & - & -
& 90.0 & 73.3 & 80.0 & 63.3
& 78.3 & 58.3 & 63.3 & 43.3
& 80.0 & 65.0 & {76.7} & 53.3
& 56.7 & 35.0 & 40.0 & 3.3 \\
& \textit{w/o}-RL
& - & - & - & -
& 100.0 & 78.3 & 81.7 & 65.0
& 73.3 & 35.0 & 68.3 & 25.0
& 85.0 & 61.7 & 80.0 & 48.3
& 55.0 & 23.3 & 43.3 & 8.3 \\
& \textit{w/o}-RL-C
& - & - & - & -
& 100.0 & 73.3 & 81.7 & 60.0
& 70.0 & 30.0 & 63.3 & 26.7
& 78.3 & 58.3 & 66.7 & 46.7
& 45.0 & 23.3 & 43.3 & 6.7 \\
\cmidrule[0.3pt](lr){2-22}
& MVoT
& - & - & - & -
& 86.7 & 46.7 & 80.0 & 56.7
& 73.3 & 15.0 & 61.7 & 3.3
& 73.3 & 56.7 & 63.3 & 46.7
& 46.7 & 13.3 & 36.7 & 3.3 \\
&  T-CoT
& - & - & - & -
& 75.0 & 33.3 & 66.7 & 33.3
& 65.0 & 5.0  & 53.3 & 0.0
& 71.7 & 53.3 & 50.0 & 36.7
& 41.7 & 6.7  & 36.7 & 0.0 \\
\midrule

\rowcolor{gray!15}
\cellcolor{white}\multirow{6}{*}{\textbf{5}}
& SVoT$_{\text{p}}$
& 93.3 & 80.0 & 70.0 & 41.7
& 91.7 & 86.7 & 85.0 & 58.3
& 78.3 & 71.7 & 73.3 & 53.3
& 93.3 & 66.7 & 83.3 & 60.0
& 46.7 & 3.3  & 33.3 & 1.7 \\
& \textit{w/o}-V
& 90.0 & 68.3 & 65.0 & 35.0
& 81.7 & 70.0 & 73.3 & 53.3
& 76.7 & 53.3 & 65.0 & 40.0
& 76.7 & 61.7 & 73.3 & 50.0
& 33.3 & 0.0 & {25.0} & 0.0 \\
& \textit{w/o}-RL
& 90.0 & 33.3 & {63.3} & 28.3
& 81.7 & 65.0 & 73.3 & 50.0
& 70.0 & 33.3 & 66.7 & 21.7
& 80.0 & 55.0 & 70.0 & 46.7
& 35.0 & 3.3  & 26.7 & 1.7 \\
& \textit{w/o}-RL-C
& 86.7 & 36.7 & 60.0 & 23.3
& 78.3 & 50.0 & 60.0 & 36.7
& 63.3 & 23.3 & {55.0} & 16.7
& 73.3 & 53.3 & 66.7 & 41.7
& 30.0 & 1.7  & 21.7 & 0.0 \\
\cmidrule[0.3pt](lr){2-22}
& MVoT
& 83.3 & 26.7 & 56.7 & 26.7
& 73.3 & 43.3 & 53.3 & 36.7
& 61.7 & 10.0 & 55.0 & 3.3
& 73.3 & 56.7 & 53.3 & 38.3
& 33.3 & 3.3  & 30.0 & 0.0 \\
&  T-CoT
& 71.7 & 16.7 & 46.7 & 16.7
& 73.3 & 33.3 & 43.3 & 26.7
& 60.0 & 3.3  & 50.0 & 0.0
& 71.7 & 51.7 & 43.3 & 31.7
& 31.7 & 0.0  & 25.0 & 0.0 \\
\midrule

\rowcolor{gray!15}
\cellcolor{white}\multirow{6}{*}{\textbf{6}}
& SVoT$_{\text{p}}$
& 90.0 & 81.7 & 66.7 & 36.7
& 93.3 & 83.3 & 76.7 & 55.0
& 71.7 & 68.3 & 65.0 & 48.3
& 83.3 & 60.0 & 75.0 & 56.7
& 45.0 & 6.7  & 30.0 & 1.7 \\
& \textit{w/o}-V
& {86.7} & 70.0 & 58.3 & 25.0
& 80.0 & 63.3 & 68.3 & 45.0
& 61.7 & 45.0 & 56.7 & 26.7
& 75.0 & 55.0 & 66.7 & 45.0
& {43.3} & 0.0 & 23.3 & 0.0 \\
& \textit{w/o}-RL
& {83.3} & 38.3 & {61.7} & 16.7
& 73.3 & 51.7 & 66.7 & 38.3
& 66.7 & 26.7 & 51.7 & 16.7
& 70.0 & 51.7 & 63.3 & 40.0
& 40.0 & 6.7  & 21.7 & 1.7 \\
& \textit{w/o}-RL-C
& {83.3} & 33.3 & 53.3 & 16.7
& 71.7 & 40.0 & 53.3 & 28.3
& 56.7 & 20.0 & 53.3 & 16.7
& 66.7 & 50.0 & 60.0 & 30.0
& {41.7} & 3.3  & 18.3 & 0.0 \\
\cmidrule[0.3pt](lr){2-22}
& MVoT
& 83.3 & 26.7 & 46.7 & 3.3
& 63.3 & 40.0 & 43.3 & 20.0
& 58.3 & 6.7  & 51.7 & 6.7
& 60.0 & 45.0 & 51.7 & 26.7
& 40.0 & 0.0  & 20.0 & 0.0 \\
&  T-CoT
& 70.0 & 16.7 & 36.7 & 1.7
& 50.0 & 36.7 & 33.3 & 13.3
& 58.3 & 3.3  & 41.7 & 3.3
& 48.3 & 36.7 & 40.0 & 28.3
& 38.3 & 0.0  & 23.3 & 0.0 \\
\midrule

\rowcolor{gray!15}
\cellcolor{white}\multirow{6}{*}{\textbf{7}}
& SVoT$_{\text{p}}$
& 80.0 & 70.0 & 65.0 & 33.3
& - & - & - & -
& - & - & - & -
& - & - & - & -
& - & - & - & - \\
& \textit{w/o}-V
& {78.3} & 50.0 & 48.3 & 21.7
& - & - & - & -
& - & - & - & -
& - & - & - & -
& - & - & - & - \\
& \textit{w/o}-RL
& {78.3} & 33.3 & {48.3} & 13.3
& - & - & - & -
& - & - & - & -
& - & - & - & -
& - & - & - & - \\
& \textit{w/o}-RL-C
& 73.3 & 26.7 & 46.7 & 6.7
& - & - & - & -
& - & - & - & -
& - & - & - & -
& - & - & - & - \\
\cmidrule[0.3pt](lr){2-22}
& MVoT
& 73.3 & 6.7  & 40.0 & 0.0
& - & - & - & -
& - & - & - & -
& - & - & - & -
& - & - & - & - \\
& T-CoT
& 60.0 & 3.3  & 30.0 & 0.0
& - & - & - & -
& - & - & - & -
& - & - & - & -
& - & - & - & - \\
\bottomrule
\end{tabular}
}
\caption{Full ablation study of SVoT$_\text{p}$ evaluated on goal coverage accuracy, using the same settings as Table \ref{tab:results_by_size}. \textit{w/o}-V denotes SVoT$_\text{p}$ without visualizations; \textit{w/o}-RL denotes SVoT$_\text{p}$ without GRPO (trained using only SFT for 10 epochs); and \textit{w/o}-RL-C represents SFT with only state representations and visualizations without transition reasoning chains.}
\label{tab:full_ablation}
\end{table*}

\newcommand{\maze}{\textsc{Maze}\xspace}

\captionsetup[table]{font=normalsize}

\begin{table*}[]
\centering
\begin{tcolorbox}[title = {\textbf{SVoT+PRM input prompts and responses in \maze}}]
\textbf{\maze} (Free Response)
\\
\\
Task: Maze \\
Determine the positions of the player following the action sequence. Please include the reasoning process, starting with \texttt{<think>} and ending with \texttt{</think>}, to derive the positions of the player and the new image after executing the actions. \\
The definitions of the actions are given below.\\
Go up / left / down / right moves the agent by one grid cell in the corresponding absolute direction, subject to the following preconditions and effects.\\
\textbf{Preconditions}: the adjacent cell in the chosen direction is outside the grid boundary. \\
\textbf{Effects}: the agent remains at its current position (out-of-bounds movement is invalid). \\
\textbf{Preconditions}: the adjacent cell in the chosen direction is a wall. \\
\textbf{Effects}: the agent remains at its current position (movement blocked by an obstacle). \\
\textbf{Preconditions}: the adjacent cell in the chosen direction is inside the grid boundary and is not a wall (i.e., a clear cell). \\
\textbf{Effects}: the agent moves one cell in the chosen direction. \\
Full Action Sequence: Go right, Go right.  \\
Coordinates are (row, column), starting at (0,0) in the top-left corner. \\
Initial maze: \textless image\textgreater \\
Initial player position: (2, 0). \\
\\
\textcolor{red}{\# Image prediction}\\
Input: \ldots Go right, player position: (2,1).\\
Response: \textit{{<think>} \ldots {</think>} \textless image\textgreater}
\\
\textcolor{red}{\# State prediction}\\
Input: \ldots Go right, player position: (2,1). \textless image\textgreater \\   
Response: \textit{{<think>} \ldots {</think>}}       \textit{Go right},   \textit{player position}:
\textit{(2,2)}.
\\
\textcolor{red}{\# Goal prediction}\\
Input: \ldots Go right, player position: (2,1). Go right, player position: (2,2). \textless image\textgreater \\
Response: \textit{Action sequence finished}. \textit{Player position}: \textit{(2,2)}.
\\
\\
\textbf{\maze} (Classification)
\\
\\
Task: Maze Navigation Simulation 
\\
...(same as free response)
\\
Initial maze: \textless image\textgreater \\
Initial player position: (2, 0). \\
\textbf{Return A, B, C or D.} \\
\textbf{A. Player position: (2,0).} \\
\textbf{B. Player position: (2,2).} \\
\textbf{C. Player position: (1,2).} \\
\textbf{D. Player position: (2,1).}  \\
\\
\textcolor{red}{\# Image prediction}\\
Response: \ldots (same as free response)
\\
\textcolor{red}{\# State prediction}\\
Response: \ldots (same as free response)
\\
\textcolor{red}{\# Goal prediction}\\
Response: \textit{Action sequence finished}. \textbf{\textit{The answer is}}: \textbf{\textit{B}}.
\end{tcolorbox}
\caption{Examples SVoT with PRM input prompts and model responses in the \maze domain under free-response and classification settings (we omit duplicated prompts and responses for conciseness). \textit{Italicized} text denotes the expected response. }
\label{apptab:svot-maze}
\end{table*}
\newcommand{\frozenlake}{\textsc{FrozenLake}\xspace}
\captionsetup[table]{font=normalsize}

\begin{table*}[]
\centering
\begin{tcolorbox}[title = {\textbf{SVoT+PRM input prompts and responses in \frozenlake}}]
\textbf{\frozenlake} (Free Response)
\\
\\
Task: FrozenLake \\
Determine the position of player and the state of player (ice, hole, or goal) following the action sequence. Please include the reasoning process, starting with \texttt{<think>} and ending with \texttt{</think>}, to derive the position of the player, the state of player, and the new image after executing the actions. \\
The definitions of the actions are given below.\\
Go up / left / down / right moves the agent by one grid cell in the corresponding absolute direction, subject to the following preconditions and effects.\\
\textbf{Preconditions}: the adjacent cell in the chosen direction is ice. \\
\textbf{Effects}: the agent moves one cell in the chosen direction. The player's state remains ice. \\
\textbf{Preconditions}: the adjacent cell in the chosen direction is a hole. \\
\textbf{Effects}: the agent moves one cell into the hole and falls in. The player's state becomes hole. \\
\textbf{Preconditions}: the adjacent cell in the chosen direction is the goal. \\
\textbf{Effects}: the agent moves into the goal cell and successfully reaches the goal. The player's state becomes goal. \\
Full Action Sequence: Go down, Go down.  \\
Coordinates are (row, column), starting at (0,0) in the top-left corner. \\
Initial maze: \textless image\textgreater \\
Initial player position: (0, 0), state: ice.
\\
\\
Input: \ldots Go down, player position: (1,0), state: ice. Go down, player position: (2,0), state: ice. \textless image\textgreater \\
Response: \textit{Action sequence finished}. \textit{player position}: \textit{(2,0)}, \textit{state}: \textit{ice}.
\end{tcolorbox}
\caption{Example SVoT with PRM input prompts and model responses in the \frozenlake domain for free-response goal prediction. \textit{Italicized} text denotes the expected response. }
\label{apptab:svot-frozenlake}
\end{table*}
\newcommand{\sokoban}{\textsc{Sokoban}\xspace}

\captionsetup[table]{font=normalsize}

\begin{table*}[]
\centering
\begin{tcolorbox}[title = {\textbf{SVoT+PRM input prompts and responses in \sokoban}}]
\textbf{\sokoban} (Free Response)
\\
\\
Task: Sokoban \\
Determine the positions of the player and block following the action sequence. Please include the reasoning process, starting with \texttt{<think>} and ending with \texttt{</think>}, to derive the positions of the player and block, and the new image after executing the actions. \\
The definitions of the actions are given below.\\
Go up / left / down / right moves the agent by one grid cell in the corresponding absolute direction, subject to the following preconditions and effects.\\
\textbf{Preconditions}: The adjacent cell in the chosen direction is a block, and the cell immediately beyond that block in the same direction is empty and within the grid. \\
\textbf{Effects}: The agent moves one cell in the chosen direction, and the block is pushed one cell in the same direction.\\
\textbf{Preconditions}: The adjacent cell in the chosen direction is empty and within the grid. \\
\textbf{Effects}: The agent moves one cell in the chosen direction, and the block positions remain unchanged. \\
\textbf{Preconditions}: The adjacent cell in the chosen direction is a wall, outside the grid, or a block whose next cell is not empty. \\
\textbf{Effects}: The agent and all blocks remain unchanged. \\
Full Action Sequence: Go up, Go left.  \\
Coordinates are (row, column), starting at (0,0) in the top-left corner. \\
Initial maze: \textless image\textgreater \\
Initial player position: (2, 3), block position(s): (2,1), (1,2).\\
\\
Input: \ldots Go up, player position: (1,3), block position(s): (2,1), (1,2). Go left, player position: (1,2), block position(s): (2,1), (1,1). \textless image\textgreater \\ 
Response: \textit{Action sequence finished}. \textit{Player position}: \textit{(1,2)}, \textit{block position(s)}: \textit{(2,1)}, \textit{(1,1)}.
\end{tcolorbox}
\caption{Example SVoT with PRM input prompts and model responses in the \sokoban domain for free-response goal
prediction. \textit{Italicized} text denotes the expected response. }
\label{apptab:svot-sokoban}
\end{table*}
\newcommand{\pacman}{\textsc{Pacman}\xspace}

\captionsetup[table]{font=normalsize}

\begin{table*}[]
\centering
\begin{tcolorbox}[title = {\textbf{SVoT+PRM input prompts and responses in \pacman}}]
\textbf{\pacman} (Free Response)
\\
\\
Task: Pacman \\
Determine the position of the player and total number of coins collected by the player along the path by following the given action sequence. Please include the reasoning process, starting with \texttt{<think>} and ending with \texttt{</think>}, to derive the position of the player, the total number of coins collected, and the new image after executing the actions. \\
The definitions of the actions are given below.\\
Go up / left / down / right moves the agent by one grid cell in the corresponding absolute direction, subject to the following preconditions and effects.\\
\textbf{Preconditions}: The adjacent cell in the chosen direction contains a coin. \\
\textbf{Effects}: The agent moves one cell in the chosen direction, and the number of collected coins increases by one. \\
\textbf{Preconditions}: The adjacent cell in the chosen direction is clear (i.e., does not contain a coin). \\
\textbf{Effects}: The agent moves one cell in the chosen direction, and the number of collected coins remains unchanged. \\
Full Action Sequence: Go up, Go up.  \\
Coordinates are (row, column), starting at (0,0) in the top-left corner. \\
Initial maze: \textless image\textgreater \\
Initial player position: (3, 3), collected coin(s): 0.
\\
\\
Input: \ldots Go up, player position: (2,3), collected coin(s): 1. Go up, player position: (1,3), collected coin(s): 2. \textless image\textgreater \\
Response:  \textit{Action sequence finished}. \textit{Player position}: \textit{(1,3)}, \textit{collected coin(s)}: \textit{2}.
\end{tcolorbox}
\caption{Example SVoT with PRM input prompts and model responses in the \pacman domain for free-response goal
prediction. \textit{Italicized} text denotes the expected response. }
\label{apptab:svot-pacman}
\end{table*}
\newcommand{\Gather}{\textsc{Gather}\xspace}

\captionsetup[table]{font=normalsize}

\begin{table*}[]
\centering
\begin{tcolorbox}[title = {\textbf{SVoT+PRM input prompts and responses in \Gather}}]
\textbf{\Gather} (Free Response)
\\
\\
\textbf{Task: Gather} \\
Determine the position of player and the number of balls of each color collected along the path by following the given action sequence. Please include the reasoning process, starting with \texttt{<think>} and ending with \texttt{</think>}, to derive the position of the player, the count of collected balls for each color, and the new image after executing the actions.
\\
The definitions of the actions are given below. \\
Go up / left / down / right with a specified step count moves the agent by that number of grid cells in the corresponding absolute direction, subject to the following preconditions and effects. \\
\textbf{Preconditions}: The next specified number of cells in the chosen direction contain balls. \\
\textbf{Effects}: The agent moves the specified number of cells in the chosen direction, collects all balls along the path (including the destination cell), and increments the collected count for each ball color accordingly. \\
\textbf{Preconditions}: The next specified number of cells in the chosen direction contain no balls (i.e., all cells are clear). \\
\textbf{Effects}: The agent moves the specified number of cells in the chosen direction, and the collected ball counts remain unchanged. \\
Full Action Sequence: Go right 2, Go down 2.  \\
Coordinates are (row, column), starting at (0,0) in the top-left corner. \\
Initial maze: \textless image\textgreater \\
Initial player position: (0, 0), collected ball(s): red: 0, green: 0.
\\
\\
Input: \ldots Go right 2, player position: (0,2), collected ball(s): red: 1, green: 0. Go down 2, player position: (2,2), collected ball(s): red: 1, green: 2. \textless image\textgreater \\
Response: \textit{Action sequence finished}. \textit{Player position}: \textit{(2,2)}, \textit{collected ball(s)}: \textit{red: 1, green: 2.}
\end{tcolorbox}
\caption{Example SVoT with PRM input prompts and model responses in the \Gather domain for free-response goal
prediction. \textit{Italicized} text denotes the expected response. }
\label{apptab:svot-gather}
\end{table*}

\subsection{Visualization Quantitative Analysis} \label{Visualization Quantitative Analysis Breakdown by Grid Size}

Table~\ref{tab:appendix_full_results_styled} provides a more fine-grained breakdown of the visual generation quality reported in Table~\ref{tab:main_averaged_results} across different grid sizes. Overall, SVoT$_\text{p}$ maintains strong visual fidelity across evaluated domains and grid sizes, and in \textsc{Sokoban} at size 6, it achieves a particularly high visual reward $r_v$ of $0.8834$, where the other methods exhibit noticeable degradation. This supports the effectiveness of transition-aware supervision in mitigating visual hallucinations in multi-object interactive environments. However, for \textsc{FrozenLake} with size 4, SVoT$_{\text{p}}$ exhibits slightly lower visual quality compared to SVoT$_{\text{o}}$. Finally, visual reward scores for all methods drop markedly in the \textsc{Gather} domain at grid size 6, highlighting clear direction for future work.

We additionally report three foreground--background correctness metrics in Table~\ref{tab:fg_bg_full_visualization_metrics}.  These metrics evaluate whether generated visualizations preserve object-level correctness in grid cells. 
Foreground Cell Accuracy measures whether cells containing ground-truth foreground objects are correctly generated, while Background Error Rate measures hallucinated foreground objects in cells that are background in the ground-truth state. 
We further report Full Visualization Accuracy, which counts a visualization as correct only when all ground-truth foreground cells are correctly generated and no ground-truth background cell contains an extra foreground object. These diagnostic metrics are enabled by our PDDL-based domain construction, which provides ground-truth cell-level object labels for each image. In broader applications, such semantic labels are often unavailable; therefore, the visual reward $r_v$, which uses a variance-based foreground heuristic, is broadly applicable.

Table~\ref{tab:fg_bg_full_visualization_metrics} shows that SVoT$_\text{p}$ consistently achieves higher Foreground Cell Accuracy and Full Visualization Accuracy than MVoT and SVoT$_\text{o}$, while also reducing Background Error Rate in most settings. 
Together with the reward-based analysis in Table~\ref{tab:appendix_full_results_styled}, these results indicate that SVoT$_\text{p}$ improves both visual fidelity and object-level correctness in generated visualizations. 

\subsection{Detailed Ablation Study}  \label{Detailed Ablation Study}

Table~\ref{tab:full_ablation} extends Table~\ref{tab:ablation_study} by additionally reporting classification results for the variants of SVoT$_\text{p}$. While classification performance across variants broadly tracks the trends observed in the free-response setting, the classification metric is noticeably less sensitive to the removal of RL in certain domains. For example, in the \textsc{Maze} domain with size 6, \textit{w/o}-RL attains high classification accuracy (83.3\%), yet its free-response score collapses dramatically (38.3\% vs.\ 81.7\%). This gap indicates that the SFT-only variant (\textit{w/o}-RL) tends to exploit shortcuts to predict the final outcome label without truly learning the underlying reasoning dynamics. By contrast, the strong alignment between classification and free-response performance in SVoT$_\text{p}$ suggests that its gains arise from genuine spatial reasoning rather than superficial pattern matching.

\subsection{Additional  Visualizations} \label{Additional Visualizations}

Figure~\ref{fig:images_app} illustrates additional visualization examples for the \textsc{Maze} and \textsc{Sokoban} domains, omitted from Figure~\ref{fig:images}. In both domains, SVoT$_\text{p}$ produces consistently high-fidelity visualizations that faithfully follow the transition reasoning chain: in \textsc{Maze}, the agent correctly moves one step to the left, and in \textsc{Sokoban}, the player accurately moves one step down and updates the block position accordingly. By contrast, SVoT$_\text{o}$ generates incorrect visualizations due to erroneous movements in all examples. As in examples in Figure~\ref{fig:images}, MVoT tends to generate blurrier and less accurate visualizations.

\subsection{Hyperparameters Analysis in Visual Reward} \label{Hyperparameters Analysis in Visual Reward}

The visual reward $r_v$ in SVoT involves several hyperparameters, including the threshold $\delta$ for foreground--background identification, the threshold $\tau$ for cell matching, the foreground weighting used to emphasize semantically salient cells, and the weights $\lambda_h$, $\lambda_e$, and $\lambda_c$ assigned to $s_{\mathrm{hash}}$, $s_{\mathrm{edge}}$, and $s_{\mathrm{color}}$, respectively, in the computation of the composite similarity score $\mathcal{S}$. Tables \ref{tab:svotp_delta}--~\ref{tab:svotp_lambda} report the results of hyperparameter study for $r_v$  in SVoT$_\text{p}$ on the validation sets of the five introduced domains with grid sizes $5$ and $6$ under the free-response setting.

Table~\ref{tab:svotp_delta} indicates that $\delta=0.002$ yields the most balanced performance across domains. While $\delta=0.001$ remains competitive, increasing the threshold to $\delta=0.003$ leads to a decline, suggesting that an overly restrictive foreground criterion may exclude semantically salient regions from the foreground mask, causing them to receive lower weights in reward computation. Table~\ref{tab:svotp_tau} indicates that stricter matching thresholds are beneficial, with both $\tau=0.8$ and $\tau=0.9$ outperforming $\tau=0.7$, while $\tau=0.8$ provides the most consistent overall performance. The results in Table~\ref{tab:svotp_fgweight} underscores the importance of foreground emphasis, as the settings that assign higher weights to foreground cells outperform the uniform baseline. The strongest results are obtained with a $10\times$ ratio, whereas increasing the ratio further to $20\times$ leads to mild degradation on several domains. Table~\ref{tab:svotp_lambda} shows that the three similarity components are most effective when used jointly. Among the single-component variants, the perceptual-hash term ($\lambda_h=1$) yields the strongest standalone performance, indicating that coarse structural correspondence provides the dominant signal for reliable cell matching, while the edge-only ($\lambda_e=1$) and color-only ($\lambda_c=1$) settings contribute complementary information. These findings are consistent with the superior performance of the combined setting $(\lambda_h,\lambda_e,\lambda_c)=(0.4,0.3,0.3)$.

We also conduct an ablation study on $r_v$ in SVoT${_\text{p}}$ to evaluate the respective contributions of $r_{\mathrm{str}}$ and $r_{\mathrm{qua}}$. The results are reported in Table~\ref{tab:svotp_reward_comp}. The term $r_{\mathrm{str}}$ provides the stronger standalone supervisory signal, whereas $r_{\mathrm{qua}}$ is less effective when used in isolation. Their combination nevertheless yields the best overall performance, indicating that both cell-level consistency, as provided by $r_{\mathrm{str}}$, and perceptual quality, as captured by $r_{\mathrm{qua}}$, are important for achieving strong performance.

\subsection{Limitations and Future Work} \label{Limitations_Section}

SVoT introduces additional inference cost compared with outcome-only or text-only baselines, since it autoregressively generates structured intermediate states and visualizations guided by transition reasoning chains along the trajectory. This overhead is a deliberate trade-off for making sequential state tracking explicit, verifiable, and reward-optimizable. In future work, the efficiency of SVoT could be further improved through selective visualization, adaptive reasoning length, and intermediate-state caching, while preserving the benefits of explicit transition modeling.

\begin{table}[t]
    \centering 
    \setlength{\tabcolsep}{4pt} 
    \renewcommand{\arraystretch}{1} 
\begin{tabular}{c !{\color{lightgray}\vrule} l !{\color{lightgray}\vrule} c !{\color{lightgray}\vrule} c !{\color{lightgray}\vrule} c !{\color{lightgray}\vrule} c !{\color{lightgray}\vrule} c}
    \toprule
    \textbf{Size} & \textbf{Setting}
    & \textbf{Maze} & \textbf{FrozenLake} & \textbf{Sokoban} & \textbf{Pacman} & \textbf{Gather} \\
    \midrule
        \multirow{3}{*}{5} 
        & $\delta=0.001$ 
        & 83.3 & \textbf{91.7} & \textbf{81.7} & \textbf{73.3} & 10.0 \\
        & \cellcolor{gray!15}$\delta=0.002$ 
        & \cellcolor{gray!15}\textbf{85.0} & \cellcolor{gray!15}\textbf{91.7} & \cellcolor{gray!15}\textbf{81.7} & \cellcolor{gray!15}\textbf{73.3} & \cellcolor{gray!15}\textbf{11.7} \\
        & $\delta=0.003$ 
        & 73.3 & 83.3 & 75.0 & 66.7 & 3.3 \\
        \midrule
        \multirow{3}{*}{6} 
        & $\delta=0.001$ 
        & 83.3 & 88.3 & 75.0 & 68.3 & \textbf{11.7} \\
        & \cellcolor{gray!15}$\delta=0.002$ 
        & \cellcolor{gray!15}\textbf{86.7} & \cellcolor{gray!15}\textbf{90.0} & \cellcolor{gray!15}\textbf{78.3} & \cellcolor{gray!15}\textbf{70.0} & \cellcolor{gray!15}{10.0} \\
        & $\delta=0.003$ 
        & 76.7 & 73.3 & 68.3 & 66.7 & 0.0 \\
        \bottomrule
    \end{tabular}
    \vspace{0.2cm}
    \caption{Parameter study on the foreground--background threshold $\delta$ for $r_v$ across five domains under the free-response setting for grid sizes $5$ and $6$. Best results are shown in \textbf{bold}, and the adopted parameter setting for $r_v$ is highlighted in gray.}
    \label{tab:svotp_delta}
\end{table}

\begin{table}[t]
    \centering
    \setlength{\tabcolsep}{4pt}   
    \renewcommand{\arraystretch}{1} 
\begin{tabular}{c !{\color{lightgray}\vrule} l !{\color{lightgray}\vrule} c !{\color{lightgray}\vrule} c !{\color{lightgray}\vrule} c !{\color{lightgray}\vrule} c !{\color{lightgray}\vrule} c}
    \toprule
    \textbf{Size} & \textbf{Setting}
    & \textbf{Maze} & \textbf{FrozenLake} & \textbf{Sokoban} & \textbf{Pacman} & \textbf{Gather} \\
    \midrule
        \multirow{3}{*}{5}
        & $\tau=0.7$ 
        & 73.3 & 80.0 & 70.0 & 60.0 & 1.7 \\
        & \cellcolor{gray!15}$\tau=0.8$ 
        & \cellcolor{gray!15}\textbf{85.0} & \cellcolor{gray!15}\textbf{91.7} & \cellcolor{gray!15}\textbf{81.7} & \cellcolor{gray!15}\textbf{73.3} & \cellcolor{gray!15}\textbf{11.7} \\
        & $\tau=0.9$ 
        & 83.3 & \textbf{91.7} & 80.0 & \textbf{73.3} & \textbf{11.7} \\
        \midrule
        \multirow{3}{*}{6}
        & $\tau=0.7$ 
        & 71.7 & 75.0 & 75.0 & 61.7 & 0.0 \\
        & \cellcolor{gray!15}$\tau=0.8$ 
        & \cellcolor{gray!15}\textbf{86.7} & \cellcolor{gray!15}\textbf{90.0} & \cellcolor{gray!15}\textbf{78.3} & \cellcolor{gray!15}\textbf{70.0} & \cellcolor{gray!15}\textbf{10.0} \\
        & $\tau=0.9$ 
        & \textbf{86.7} & \textbf{90.0} & 75.0 & \textbf{70.0} & 6.7 \\
        \bottomrule
    \end{tabular}
    \vspace{0.2cm}
    \caption{Parameter study on the cell-matching threshold $\tau$ for $r_v$ across five domains under the free-response setting for grid sizes $5$ and $6$. Best results are shown in \textbf{bold}, and the adopted parameter setting for $r_v$is highlighted in gray.}
    \label{tab:svotp_tau}
\end{table}

\begin{table}[t]
    \centering
    \setlength{\tabcolsep}{4pt}  
    \renewcommand{\arraystretch}{1} 

\begin{tabular}{c !{\color{lightgray}\vrule} l !{\color{lightgray}\vrule} c !{\color{lightgray}\vrule} c !{\color{lightgray}\vrule} c !{\color{lightgray}\vrule} c !{\color{lightgray}\vrule} c}
    \toprule
    \textbf{Size} & \textbf{Setting}
    & \textbf{Maze} & \textbf{FrozenLake} & \textbf{Sokoban} & \textbf{Pacman} & \textbf{Gather} \\
    \midrule
        \multirow{3}{*}{5}
        & $1\text{x}$ 
        & 73.3 & 85.0 & 71.7 & 70.0 & 0.0 \\
        & \cellcolor{gray!15}$10\text{x}$ 
        & \cellcolor{gray!15}\textbf{85.0} & \cellcolor{gray!15}\textbf{91.7} & \cellcolor{gray!15}\textbf{81.7} & \cellcolor{gray!15}\textbf{73.3} & \cellcolor{gray!15}\textbf{11.7} \\
        & $20\text{x}$  
        & 80.0 & 85.0 & 73.3 & \textbf{73.3} & 8.3 \\
        \midrule
        \multirow{3}{*}{6}
        & $1\text{x}$ 
        & 78.3 & 81.7 & 68.3 & 65.0 & 0.0 \\
        & \cellcolor{gray!15}$10\text{x}$ 
        & \cellcolor{gray!15}\textbf{86.7} & \cellcolor{gray!15}\textbf{90.0} & \cellcolor{gray!15}\textbf{78.3} & \cellcolor{gray!15}{70.0} & \cellcolor{gray!15}{10.0} \\
        & $20\text{x}$  
        & 81.7 & 83.3 & 71.7 & \textbf{73.3} & \textbf{11.7} \\
        \bottomrule
    \end{tabular}
    \vspace{0.2cm}
    \caption{Parameter study on the foreground weighting ratio for $r_v$ across five domains under the free-response setting for grid sizes $5$ and $6$. Best results are shown in \textbf{bold}, and the adopted parameter setting for $r_v$ is highlighted in gray.}
    \vspace{0.2cm}
    \label{tab:svotp_fgweight}
\end{table}

\begin{table}[t]
    \centering
    \setlength{\tabcolsep}{4pt} 
    \renewcommand{\arraystretch}{1} 

\begin{tabular}{c !{\color{lightgray}\vrule} l !{\color{lightgray}\vrule} c !{\color{lightgray}\vrule} c !{\color{lightgray}\vrule} c !{\color{lightgray}\vrule} c !{\color{lightgray}\vrule} c}
    \toprule
    \textbf{Size} & \textbf{Setting}
    & \textbf{Maze} & \textbf{FrozenLake} & \textbf{Sokoban} & \textbf{Pacman} & \textbf{Gather} \\
    \midrule
        \multirow{4}{*}{5}
        & $\lambda_h = 1$, $\lambda_e = 0$, and $\lambda_c = 0$ 
        & 83.3 & 90.0 & 78.3 & 68.3 & \textbf{11.7} \\
        & $\lambda_h = 0$, $\lambda_e = 1$, and $\lambda_c = 0$  
        & 80.0 & 83.3 & 71.7 & 68.3 & 5.0 \\
        & $\lambda_h = 0$, $\lambda_e = 0$, and $\lambda_c = 1$ 
        & 73.3 & 75.0 & 66.7 & 61.7 & 0.0 \\
        & \cellcolor{gray!15}$\lambda_h = 0.4$, $\lambda_e = 0.3$, and $\lambda_c = 0.3$ 
        & \cellcolor{gray!15}\textbf{85.0} & \cellcolor{gray!15}\textbf{91.7} & \cellcolor{gray!15}\textbf{81.7} & \cellcolor{gray!15}\textbf{73.3} & \cellcolor{gray!15}\textbf{11.7} \\
        \midrule
        \multirow{4}{*}{6}
        & $\lambda_h = 1$, $\lambda_e = 0$, and $\lambda_c = 0$ 
        & 81.7 & 86.7 & 75.0 & \textbf{70.0} & 8.3 \\
        & $\lambda_h = 0$, $\lambda_e = 1$, and $\lambda_c = 0$  
        & 75.0 & 83.3 & 71.7 & {63.3} & 5.0 \\
        & $\lambda_h = 0$, $\lambda_e = 0$, and $\lambda_c = 1$ 
        & 73.3 & 71.7 & 61.7 & 58.3 & 0.0 \\
        & \cellcolor{gray!15}$\lambda_h = 0.4$, $\lambda_e = 0.3$, and $\lambda_c = 0.3$ 
        & \cellcolor{gray!15}\textbf{86.7} & \cellcolor{gray!15}\textbf{90.0} & \cellcolor{gray!15}\textbf{78.3} & \cellcolor{gray!15}\textbf{70.0} & \cellcolor{gray!15}\textbf{10.0} \\
        \bottomrule
    \end{tabular}
    \vspace{0.2cm}
    \caption{Parameter study on the component weights $(\lambda_h,\lambda_e,\lambda_c)$ in the composite similarity score $\mathcal{S}$ for $r_v$ across five domains under the free-response setting for grid sizes $5$ and $6$. Best results are shown in \textbf{bold}, and the adopted parameter setting for $r_v$ is highlighted in gray.}
    \label{tab:svotp_lambda}
\end{table}

\begin{table}[t]
    \centering
    \setlength{\tabcolsep}{4pt} 
    \renewcommand{\arraystretch}{1} 

\begin{tabular}{c !{\color{lightgray}\vrule} l !{\color{lightgray}\vrule} c !{\color{lightgray}\vrule} c !{\color{lightgray}\vrule} c !{\color{lightgray}\vrule} c !{\color{lightgray}\vrule} c}
    \toprule
    \textbf{Size} & \textbf{Setting}
    & \textbf{Maze} & \textbf{FrozenLake} & \textbf{Sokoban} & \textbf{Pacman} & \textbf{Gather} \\
    \midrule
        \multirow{3}{*}{5}
        & $r_v = r_{\mathrm{str}}$  
        & 83.3 & 90.0 & 71.7 & 66.7 & 6.7 \\
        & $r_v = r_{\mathrm{qua}}$ 
        & 71.7 & 83.3 & 65.0 & 61.7 & 0.0 \\
        & \cellcolor{gray!15}$r_v = r_{\mathrm{str}} \cdot r_{\mathrm{qua}}$ 
        & \cellcolor{gray!15}\textbf{85.0} & \cellcolor{gray!15}\textbf{91.7} & \cellcolor{gray!15}\textbf{81.7} & \cellcolor{gray!15}\textbf{73.3} & \cellcolor{gray!15}\textbf{11.7} \\
        \midrule
        \multirow{3}{*}{6}
        & $r_v = r_{\mathrm{str}}$  
        & 81.7 & \textbf{90.0} & 71.7 & 63.3 & \textbf{11.7} \\
        & $r_v = r_{\mathrm{qua}}$ 
        & 73.3 & 81.7 & 61.7 & 58.3 & 3.3 \\
        & \cellcolor{gray!15}$r_v = r_{\mathrm{str}} \cdot r_{\mathrm{qua}}$ 
        & \cellcolor{gray!15}\textbf{86.7} & \cellcolor{gray!15}\textbf{90.0} & \cellcolor{gray!15}\textbf{78.3} & \cellcolor{gray!15}\textbf{70.0} & \cellcolor{gray!15}{10.0} \\
        \bottomrule
    \end{tabular}
    \vspace{0.2cm}
    \caption{Ablation study of the visual reward $r_v$ in SVoT$_\text{p}$ on the validation sets across five domains under the free-response setting for grid sizes $5$ and $6$. Best results are shown in \textbf{bold}, and the adopted reward composition for $r_v$ is highlighted in gray.}
    \label{tab:svotp_reward_comp}
\end{table}

\end{document}